\documentclass[10pt,twocolumn,letterpaper]{article}
\usepackage[pagenumbers]{cvpr}
\usepackage{times,graphicx,amsmath,amssymb,booktabs,multirow,array,capt-of}
\usepackage{dcolumn}
\newcolumntype{N}{D{.}{.}{2.1}}
\newcolumntype{J}{D{.}{.}{3.0}}
\newcolumntype{T}{D{.}{.}{2.2}}
\newcommand{\colhead}[1]{\multicolumn{1}{c}{#1}}
\newsavebox{\robotresultbox}

\makeatletter
\newcolumntype{B}{>{\boldmath\DC@{.}{.}{2.1}}c<{\DC@end}}
\makeatother
\newcommand{\best}[1]{\multicolumn{1}{B}{#1}}
\newcommand{\bestlast}[1]{\multicolumn{1}{B@{}}{#1}}

\usepackage{xcolor}
\definecolor{cvprblue}{rgb}{0.21,0.49,0.74}
\ifdefined\SubmissionOnly
\usepackage{xr-hyper}
\newcommand{\suppref}[1]{\ref*{supp-#1}}
\else
\newcommand{\suppref}[1]{\ref{#1}}
\fi
\usepackage[pagebackref,breaklinks,colorlinks,allcolors=cvprblue]{hyperref}
\ifdefined\SubmissionOnly\fi
\def\confName{CVPR}\def\confYear{2027}

\newcommand{\R}{\mathbb{R}}

\newcommand{\norm}[1]{\left\lVert#1\right\rVert}
\DeclareMathOperator{\Log}{Log}
\DeclareMathOperator{\Exp}{Exp}
\title{Learning from Incomplete Actions: Interval Motion Supervision for Vision-Language-Action Policies}
\author{Yunpeng Mei\thanks{Equal contribution.}\quad Chenyu Wang\footnotemark[1]\quad Xiaowen Zhu\quad Chenbo Xin\\
Hongjie Cao\quad Yiyuan Li\quad Chengye Zhang\quad YuAn Wang\quad Gang Wang\thanks{Corresponding author.}\\[0.3em]
Beijing Institute of Technology}
\begin{document}
\maketitle
\begin{abstract}
Visual labeling of egocentric demonstrations often leaves contiguous gaps in action supervision. Reliable poses around a gap still provide its net translation and rotation. We use this information to train vision-language-action policies while preserving ordinary flow matching on reliable local actions. A composition objective constrains the aggregate motion of missing-action estimates, combining multiple fixed windows when a gap exceeds the policy horizon. A shared interval branch learns from reliable endpoints and supplies the composition target. On LIBERO with $20\%$ of demonstrations and $70.5\%$ missing actions, the main method achieves $87.9\%$ mean success over three training runs. In one paired comparison with matched observation exposure and training cost, its advantage over masking is $10.5\%$. On Ego-informed RoboTwin gaps, success is $45.4\%$, versus $40.3\%$ for masking. On five real-robot tasks, mean success is $59.6\%$, versus $54.8\%$ for interval flow matching alone and $48.8\%$ for masking.
\end{abstract}

\section{Introduction}
\begin{figure*}[!t]
\centering\includegraphics[width=\textwidth]{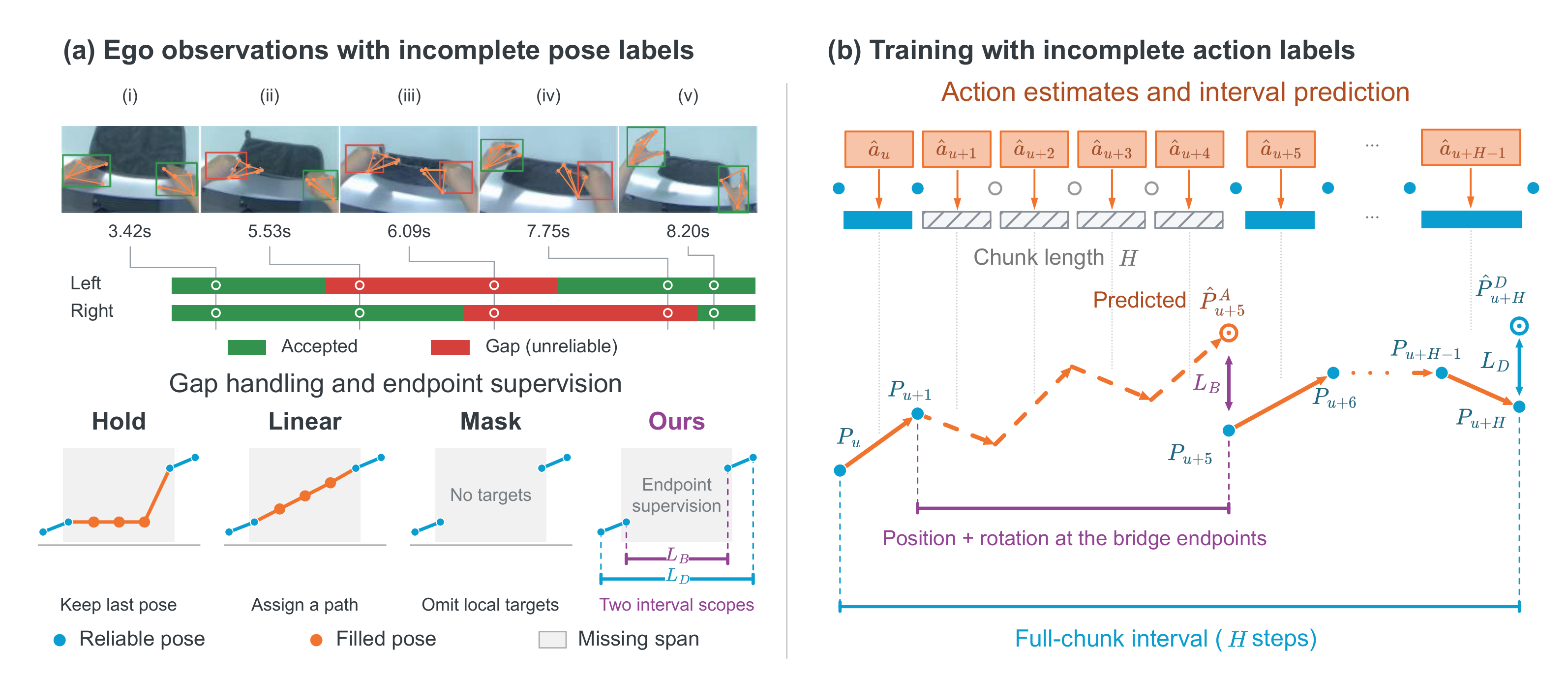}
\caption{\textbf{Training from local labels and observed endpoint motion.} (a) Illustrative pose offsets and availability tracks on real Ego frames compare Hold, Linear, Mask, and endpoint supervision. (b) A chunk starts at step $u$ and contains $H$ action estimates $\widehat a_i$. Solid blue points are observed poses $P_i$. Rings are endpoint predictions from composed actions ($\widehat P^A$) or interval prediction ($\widehat P^D$). Dashed steps lack local targets. Three missing poses remove four action targets. Composition loss $L_B$ constrains the inner bridge, and interval FM loss $L_D$ supervises the outer span.}
\label{fig:principle}
\label{fig:ego-gap-illustration}
\end{figure*}

Egocentric demonstrations can supplement robot teleoperation~\cite{egomimic}, but visual motion labels can become unreliable under occlusion or tracking failure. HOT3D benchmarks this hand and object tracking problem~\cite{hot3d}. Filtering unreliable estimates leaves contiguous gaps in action supervision. Figure~\ref{fig:ego-gap-illustration}(a) illustrates how these gaps can affect either hand independently while visual observations remain available, using constructed pose offsets on real Ego frames. How can a vision-language-action (VLA) policy~\cite{pi05} learn from partial action labels and partly observed interval motion?

We study end-effector pose increments: the policy learns stepwise translation and rotation, while reliable poses around a gap specify their net change. Mapped Ego trajectories and robot demonstrations share this supervision structure. Each interval requires reliable endpoints in a common frame and its starting observation.

Common treatments use only part of this supervision structure. Holding the last pose creates zero motion followed by a reacquisition jump. Interpolation assigns pointwise targets along an assumed interior path. Masking removes unreliable targets, but also leaves the known net motion unused. We instead retain local supervision wherever labels are reliable and use endpoint motion to constrain the missing segment as a whole.

The training objective follows the available information at two temporal scales (Figure~\ref{fig:principle}(b)). Reliable local actions keep their original flow-matching (FM) targets. Within a gap, we compose estimated pose increments in temporal order and constrain the resulting net motion. This leaves the individual missing actions free to depend on the observation and the pretrained policy. Long gaps share a constraint across fixed windows, each with its own observation.

We implement interval supervision with a shared action--interval expert. Its interval branch learns net motion from reliable endpoints and supplies features to action generation. Our main method uses a stopped-gradient interval prediction as the composition target, with a linear reference in the missing-action inputs. The missing actions receive aggregate supervision rather than pointwise interpolation targets. Observed endpoints provide comparison targets.

Our contribution is a training formulation that connects reliable local supervision to interval motion, including intervals that span multiple policy windows. Controlled LIBERO experiments test composition, interval prediction, and increasing gap severity. RoboTwin transfers bimanual missingness patterns from Ego data. A five-task real-robot comparison evaluates learning from robot and visually labeled Ego demonstrations with natural labeling gaps.

\section{Problem Formulation}
\subsection{Partial actions and reliable endpoints}
At episode step $t$, let $I_t$ denote camera inputs, $s_t$ the policy state, and $\ell$ a task instruction. Together they form the observation $C_t=(I_t,s_t,\ell)$. For the single-arm LIBERO formulation, a physical robot action has seven coordinates: three translation commands, three rotation commands, and a gripper command. We retain the pretrained action interface with $d_a$ padded coordinates and a fixed horizon $H=10$. Writing the normalized, padded action at step $t$ as $a_t$, a chunk is $A_t=(a_t,\ldots,a_{t+H-1})$.

Let $M_{t,i}\in\{0,1\}$ indicate whether the action at offset $i$ is reliable. In LIBERO, a gap removes the entire seven-dimensional action label, including the gripper. The bimanual RoboTwin study masks six pose coordinates per arm and retains recorded gripper labels. Missing steps receive no pointwise supervision. We distinguish excluding their tokens from attention from retaining tokens with a specified input distribution. In LIBERO world axes, let $p_t\in\R^3$ and $R_t\in SO(3)$ denote position and rotation at step $t$, where $SO(3)$ is the 3D rotation group. Reliable endpoints at steps $u$ and $v$ are $P_u=(p_u,R_u)$ and $P_v=(p_v,R_v)$, separated by $L=v-u$ actions. The principal logarithm $\Log$ maps a rotation to a skew-symmetric matrix, and $\vee$ converts it to a rotation vector. Their interval motion is
\begin{equation}
D_{u,L}=\begin{bmatrix}p_v-p_u\\ \Log(R_vR_u^\top)^\vee\end{bmatrix}\in\R^6.
\label{eq:interval}
\end{equation}
Equation~\eqref{eq:interval} concatenates translation with the rotation vector of the relative rotation. Rotation composition preserves temporal order. The target has six motion coordinates. Reliable gripper commands receive local supervision.

Each record contains reliable actions, their mask, and valid $(u,L,D_{u,L})$ intervals. The range code $q=(0,L)$ specifies zero start offset and duration $L$. Intervals read $C_u$, loading it directly if $u$ is absent from ordinary sampling.

Endpoint motion constrains net pose change while leaving interior paths, contact timing, and missing gripper commands ambiguous. The pretrained policy combines this constraint with local labels and visual context.

\section{Method}
\begin{figure*}[!t]
\centering\includegraphics[width=\textwidth]{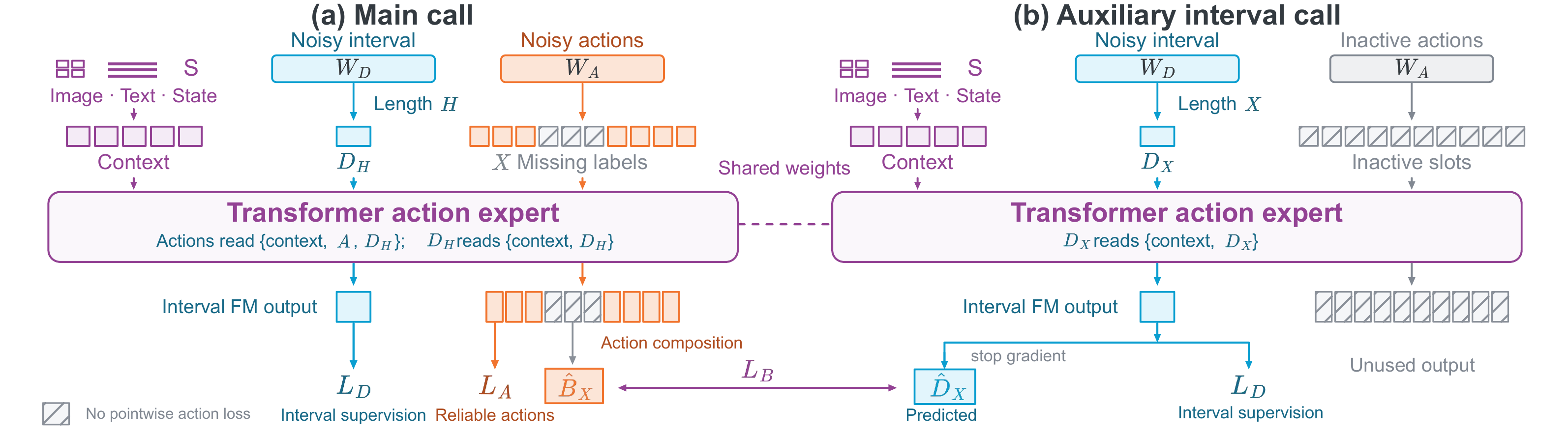}
\caption{\textbf{Shared action-interval expert.} The two LIBERO calls share weights and read interval-start observations. Squares denote tokens or readouts, embedded by $W_A$ and $W_D$. (a) $L_A$ supervises reliable actions. Hatched actions receive composition supervision through $L_B$, comparing $\widehat B_X$ with the stopped-gradient prediction $\widehat D_X$. (b) Auxiliary calls predict variable-length intervals with unused action outputs. Both train $L_D$. Actions read context, actions, and $D_H$. Intervals read only context and themselves. Inference uses the main call.}
\label{fig:architecture}
\end{figure*}

Three objectives connect local and interval supervision (Figure~\ref{fig:architecture}): $L_A$ supervises reliable actions, $L_D$ learns interval motion, and $L_B$ constrains composed action estimates across one or more windows. The main method uses a stopped-gradient interval prediction as its composition target. A comparison variant uses observed endpoints directly.

\subsection{Reliable-action flow matching}
Reliable labels remain useful even when the same window overlaps a gap. Let $\tau\in[0,1]$ denote flow time, with data at zero and noise at one. Within a window, $i$ indexes action slots. For independent standard Gaussian noise $\epsilon_i$, define the noised input $x_{\tau,i}$ and action velocity target $u_i^A$ as
\begin{equation}
 x_{\tau,i}=(1-\tau)a_i+\tau\epsilon_i,\qquad
 u^A_i=\epsilon_i-a_i.
\end{equation}
The model parameters $\theta$ define the action velocity prediction $v^A_\theta$. Using $b$ for batch rows, $i$ for action offsets, and $\|\cdot\|$ for the Euclidean norm, the action objective is
\begin{equation}
 L_A=\frac{\sum_{b,i:\,M_{b,i}=1}\norm{v^A_{\theta,b,i}-u^A_{b,i}}^2/d_a}
 {\max(\sum_{b,i}M_{b,i},1)}.
\label{eq:action}
\end{equation}
Reliable rows retain the base FM targets and padded-coordinate reduction. Reliable action predictions also attend to the missing-position inputs in Section~\ref{sec:missing-inputs}. For $M_{b,i}=0$, pointwise action FM has zero weight. Composition constrains the six motion coordinates in aggregate, leaving missing gripper commands unsupervised.

\subsection{Flow matching on observable interval motion}
Reliable endpoints provide interval targets even when their interior action labels are missing. An ordinary main view predicts the $H$-step interval from its starting observation. An auxiliary view predicts a variable-length interval from that interval's own starting observation. Write $D$ for interval motion with its start index omitted. The duration-dependent normalizer $\mathcal N_L$ produces target $z=\mathcal N_L(D)$. Normalization uses frozen action statistics and interval duration (supplementary Sec.~\suppref{sec:supp-recipe}).
For independent standard Gaussian interval noise $\eta\in\R^6$, define the noised interval $d_\tau$ and velocity target $u^D$ as
\begin{equation}
 d_\tau=(1-\tau)z+\tau\eta,\qquad u^D=\eta-z.
\end{equation}
Let $v^D_\theta$ be the predicted interval velocity. Index intervals by $j$ within batch row $b$, with $V_{b,j}\in\{0,1\}$ indicating valid endpoint supervision. Pooling both interval types gives
\begin{equation}
 L_D=\frac{\sum_{b,j:\,V_{b,j}=1}\norm{v^D_{\theta,b,j}-u^D_{b,j}}^2/6}
 {\max(\sum_{b,j}V_{b,j},1)}.
\label{eq:interval_loss}
\end{equation}
Both views share one update and unit interval-loss weight.

\subsection{Endpoint constraints on action composition}
The composition objective transfers interval supervision to the action outputs inside a gap. Denote the single-time action and interval estimates by $\widehat a_i$ and $\widehat D$, respectively. Using the inverse normalizer $\mathcal N_L^{-1}$, one velocity evaluation gives
\begin{equation}
\widehat a_i=x_{\tau,i}-\tau v^A_{\theta,i},\qquad
\widehat D=\mathcal N_L^{-1}(d_\tau-\tau v^D_\theta).
\end{equation}
At deployment, the policy integrates the velocity field over ten Euler steps. On labeled paths, $x_\tau-\tau(\epsilon-a)=a$. Endpoint supervision constrains missing-position proxies through the input distribution. These estimates use the sampled training time, so composition needs no rollout through the ten-step sampler.

Re-index grouped estimates by episode time. For action indices $I=\{u,\ldots,v-1\}$, undo normalization of $\widehat a_i$ to obtain command-space estimates $\widehat c_i$. Superscripts $p,R$ select translation and rotation commands. Let $\Exp$ map a rotation vector to its rotation matrix. The composed translation $\widehat p_I$ and rotation $\widehat R_I$ are
\begin{equation}
\widehat p_I=0.05\sum_{i\in I}\widehat c_i^p,\quad
\widehat R_I=\Exp(0.5\widehat c_{v-1}^R)\cdots\Exp(0.5\widehat c_u^R).
\end{equation}
For duration $X=|I|$, the composed six-vector $\widehat B_X$ concatenates $\widehat p_I$ and $\Log(\widehat R_I)^\vee$. The interval-head estimate $\widehat D_X$ uses the same representation. Both omit the start index. Express the target as translation $p_I^*$ and rotation matrix $R_I^*$. With frozen translation-action scale $s_p$ and selected intervals $\mathcal I$, define residuals $r_I^p,r_I^R$ and composition loss $L_B$:
\begin{align}
r_I^p&=\frac{\widehat p_I-p_I^*}{0.05|I|s_p},\quad
r_I^R=\frac{\Log(\widehat R_I(R_I^*)^\top)^\vee}{\pi},\\
L_B&=\frac{1}{\max(|\mathcal I|,1)}\sum_{I\in\mathcal I}
\frac{\norm{r_I^p}^2+\norm{r_I^R}^2}{6}.
\label{eq:composition}
\end{align}
An empty interval set contributes zero. With composition weight $\lambda_B$, the total loss is $\mathcal L=L_A+L_D+\lambda_B L_B$. The operator $\operatorname{stopgrad}$ preserves its input value and blocks its gradient. The main target is $\operatorname{stopgrad}(\widehat D_X)$, with its rotation vector converted to a matrix. Thus $L_D$ trains interval predictions, while $L_B$ trains action estimates, including through interval features read by actions. The interval target is denoised from a noised observed interval rather than generated by a pure-noise rollout. The predicted-target method uses $\lambda_B=1$ on LIBERO and $0.1$ on RoboTwin. The LIBERO observed-target variant uses $\lambda_B=1$. We also study $\lambda_B=0$. Simulation endpoints come from permitted samples of a command-integrated trajectory. Composition-only uses observed targets without an interval branch.

\begin{figure}[!t]
\centering\includegraphics[width=\columnwidth]{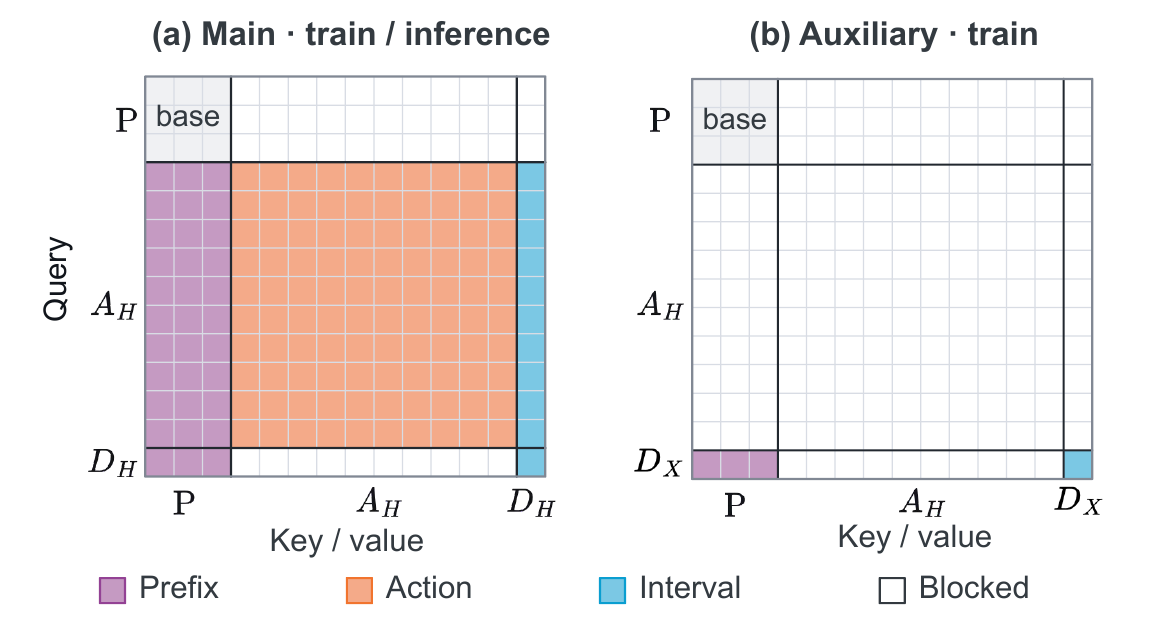}
\caption{\textbf{Attention connectivity.} Rows are queries and columns are keys/values. $P$ denotes context and $A_H$ denotes actions. Colors allow attention, white blocks it, and gray inherits prefix masking. (a) Actions read context, actions, and valid $D_H$. Inference activates $D_H$. (b) Intervals read only context and themselves. Inactive auxiliary action slots are retained for shape compatibility and do not participate in interval attention.}
\label{fig:attention}
\end{figure}

\subsection{One interval constraint across fixed windows}
\label{sec:grouped-windows}
The supervision interval can grow beyond the policy horizon while the action windows keep their original length. Ordinary sampling keeps fixed ten-action (H10) windows, including those starting or ending inside a gap. Reliable actions remain supervised. The main $D_H$ target requires reliable endpoints of that $H$-step interval.

\textbf{No gap.} We supervise all local actions, valid $D_H$ targets, and duration-conditioned auxiliary motion.

\textbf{Short gap.} $D_H$ may span a gap if both window endpoints are reliable. Otherwise its interval loss is absent while reliable-action losses remain. A complete-bridge auxiliary interval uses its exact left observation.

\begin{figure*}[!t]
\centering\includegraphics[width=\textwidth]{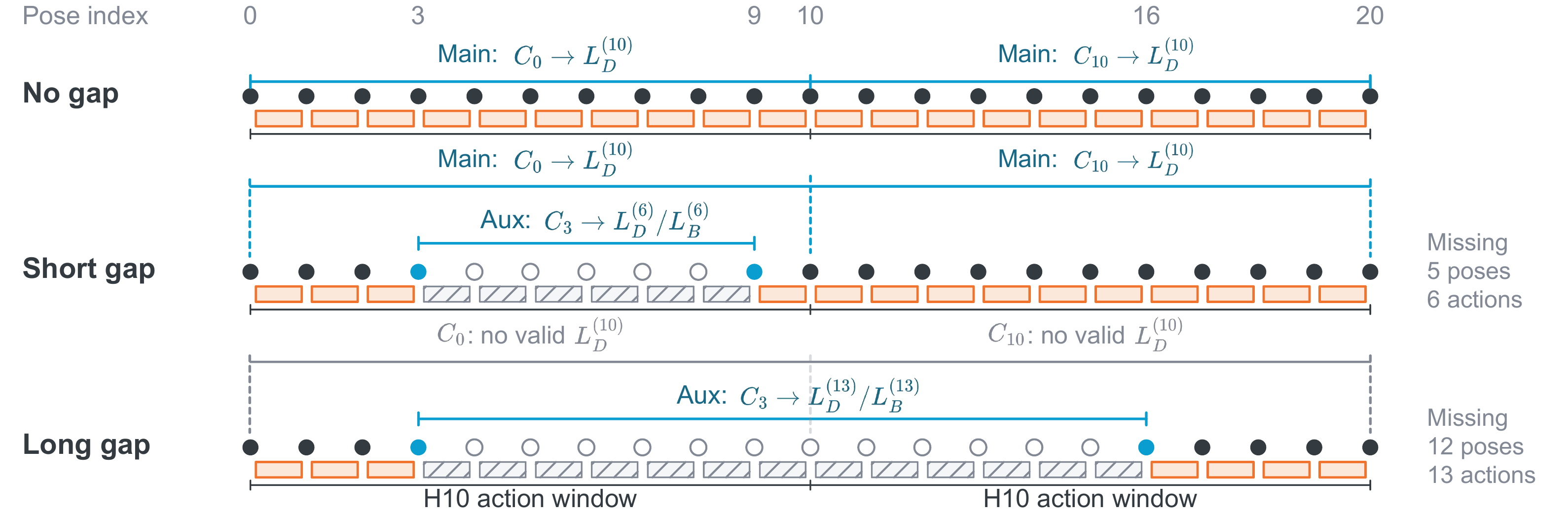}
\caption{\textbf{Fixed action windows, variable supervision intervals.} Circles denote poses, rectangles denote actions, and open circles/hatching mark missing labels. With $H=10$, five or twelve missing poses yield $X=6$ or $13$ action spans. Cyan spans mark interval FM $L_D^{(X)}$ and bridge-only composition $L_B^{(X)}$. Interval FM uses the exact start observation. Composition joins views with their own observations under one constraint. The long gap permits auxiliary $L_D^{(13)}$ but neither main $L_D^{(10)}$. H10-first prioritizes valid main intervals, then complete bridges.}
\label{fig:windows}
\end{figure*}

\textbf{Long gap.} Twelve missing poses yield a 13-action span. For a bridge $[103,116)$, the auxiliary view predicts $D_{13}$ from $C_{103}$. Windows $[100,110)$ and $[110,120)$ supply offsets $3$--$9$ and $0$--$5$ for $L_B$, respectively. Each window reads its own starting observation. Depending on alignment, two or three H10 views cover the bridge. Figure~\ref{fig:windows} contrasts fixed action windows with longer supervision intervals.

Grouped chunks use one sampled time, independent noise, and one computation graph. Their shared endpoint loss contributes to a single optimizer update. Extra grouped rows contribute only composition loss. Supplementary Sec.~\suppref{sec:supp-recipe} gives implementation details.

\subsection{Shared model and attention}
Actions and intervals use separate tokens in a shared pretrained action expert (Figure~\ref{fig:architecture}). Pretrained action projections are retained, while new interval projections embed noised motion and its duration. Tokens $D_H$ and $D_X$ predict interval velocities for their respective durations. The main view pairs actions with $D_H$ at $C_t$. Auxiliary forwards share weights and use the exact start observation $C_u$. Supplementary Sec.~\suppref{sec:supp-recipe} gives tensor shapes and embeddings.

\paragraph{Missing-action inputs.}
\label{sec:missing-inputs}
We compare three inputs for missing-action tokens in normalized action space. With Gaussian noise $\epsilon_i$, zero reference uses $x_{\tau,i}=\tau\epsilon_i$, and pure noise uses $x_{\tau,i}=\epsilon_i$. For linear reference, endpoint interpolation with a normalized-zero gripper yields $\widetilde a_i$. Its input is $x_{\tau,i}=(1-\tau)\widetilde a_i+\tau\epsilon_i$. Zero reference denotes zero in normalized action space. Pointwise action FM supervises reliable positions, whose inputs retain the ordinary FM path. All three keep missing action keys readable, so their context affects reliable predictions.

Figure~\ref{fig:attention} shows the attention mask. Action queries read the observation prefix, all action keys, and a valid main $D_H$ with $q=(0,H)$. Interval queries read only their own prefix and themselves. Auxiliary interval calls provide no action features. The interval-FM-only baseline instead blocks missing action keys and zeroes their inputs.

\subsection{Joint inference at a fixed horizon}
At deployment, independent Gaussian action and interval states are jointly denoised over ten Euler steps with $q=(0,H)$. All main tokens are active, and actions read interval features at each step. The integration-step count matches the LIBERO baseline (supplementary Sec.~\suppref{sec:supp-recipe}). Deployment requires only the current observation, with variable-length auxiliary calls confined to training.
\section{Experimental Protocol and Results}
\subsection{Data, gap construction, and training}
We use a fixed 20\% subset of the selected LIBERO suites~\cite{libero}, comprising 339 demonstrations across 40 tasks. All methods use the same visual observations and ordinary action-window sampling list. We fine-tune the same generic $\pi_{0.5}$ initialization.

Controlled gaps remove interior samples from a command-integrated pose trajectory, retain designated endpoints, and mark affected action differences as missing. LIBERO uses world-axis translation and left-multiplied rotation increments, with scales $0.05$ and $0.5$. RoboTwin uses local-frame composition in the rigid-motion group $SE(3)$. Supplementary Sec.~\suppref{sec:supp-recipe} gives the integration equations. Section~\ref{sec:real} evaluates visually labeled Ego data.

The main composition study uses 1.2-second pose gaps: twelve unavailable poses give thirteen affected action differences. Gap frequency is set to yield $70.5\%$ missing actions. The event manifest is fixed across methods.

LIBERO uses $5,000$ updates with $256$ ordinary action windows per update. Supplementary Sec.~\suppref{sec:supp-recipe} details the optimizer, learning-rate schedule, and implementation.

\subsection{Compared methods and information access}
\textbf{Hold} repeats the last reliable pose before computing labels, including rotation. Its FM input and target both use the resulting actions. \textbf{Linear} interpolates translation and rotation between endpoints, then uses the completed actions as both FM inputs and targets. \textbf{Mask} excludes missing keys and losses. \textbf{Interval FM only} adds main and auxiliary interval prediction to this blocked-key baseline. The \textbf{composition variants} retain missing keys and cross the three inputs in Section~\ref{sec:missing-inputs} with four constraint choices. Linear reference supplies a missing-action FM input without a pointwise target. Unless otherwise stated, \textbf{Ours} in the main performance comparisons denotes linear-reference inputs with predicted-interval composition over bridges. Observed-D and D-free variants isolate its components.

The base comparisons share demonstrations, ordinary sampling, gap manifests, initialization, and evaluator. Auxiliary interval views and grouped action views change exposure and computation. Section~\ref{sec:controls} specifies these controls.

\subsection{Evaluation and interval-only reference}
Each simulation model is evaluated on $800$ episodes with shared task and initialization keys across methods. Tables~\ref{tab:matrix}-\ref{tab:robotwin-planned} and supplementary Table~\suppref{tab:no-d} report one training run per setting. Supplementary Tables~\suppref{tab:supp-main} and~\suppref{tab:probes} report means over two independent training runs, and supplementary Table~\suppref{tab:predicted-seeds} reports means over three.

See supplementary Tables~\suppref{tab:supp-setup}--\suppref{tab:supp-main} for interval-only FM with roughly $30\%$ missing actions and $0.5/1.2$-second gaps.

\subsection{Supervising aggregate motion across gaps}
The high-missingness study separates the input at missing positions from the supervision applied to their aggregate motion. All cells in Table~\ref{tab:matrix} retain $L_A+L_D$ and readable missing keys. The columns add no composition loss, a stopped-gradient predicted interval, observed endpoint motion over bridges only, or observed endpoint motion with H10-first selection. ``Observed'' refers to the reliable command-space endpoints retained after label removal.

We test two observed-target scopes. \textbf{Bridge-only} constrains complete missing intervals. \textbf{H10-first} constrains each ordinary H10 with valid endpoints. If either endpoint is invalid, it collects the complete bridges overlapping that window. H10 and bridge terms share one interval-count denominator, so this change alters both interval coverage and the relative weight of bridge supervision.

\begin{table}[!ht]
\centering\small\setlength{\tabcolsep}{2pt}\renewcommand{\arraystretch}{1.08}
\caption{\textbf{Action-composition study at 70.5\% missing actions.} LIBERO, 20\% demonstrations, 1.2-second pose gaps. Success (\%). Gap input specifies the input at missing action positions. Every cell retains interval FM and readable missing keys. The ``None'' column uses $L_A+L_D$. Predicted $D$ uses a stopped-gradient single-time interval estimate as the composition target. Bridge-only uses observed bridge endpoints. H10-first prefers valid H10 endpoints, then complete overlapping bridges.}
\label{tab:matrix}
\begin{tabular*}{\columnwidth}{@{\extracolsep{\fill}}lNNNN@{}}\toprule
\multirow{2}{*}{Gap input} & \multicolumn{1}{c}{\multirow{2}{*}{None}} & \multicolumn{1}{c}{\multirow{2}{*}{Predicted $D$}} & \multicolumn{2}{c@{}}{Observed $D$}\\\cmidrule(lr){4-5}
 & & & \colhead{Bridge-only} & \multicolumn{1}{c@{}}{H10-first}\\\midrule
Zero reference & 80.3 & 87.1 & \best{87.8} & 85.4\\
Pure noise & 80.4 & 86.8 & \best{87.4} & 87.0\\
Linear reference & 79.3 & \best{88.4} & 85.3 & 85.9\\
\bottomrule
\end{tabular*}
\end{table}

\paragraph{Composition improves control across missing inputs.}
Across zero, noise, and linear inputs, observed bridge targets improve success over no composition by $6.0\%$--$7.5\%$, and predicted targets by $6.4\%$--$9.1\%$. The model with linear-reference inputs and predicted-interval composition reaches $88.4\%$, compared with $76.4\%$ for blocked-key Mask, $78.6\%$ for blocked-key interval FM, and $92.4\%$ for Clean. The $12.0\%$ gain over Mask includes changes in attention, auxiliary observations, and composition. Each row fixes the input and interval branch.

\paragraph{Target choice interacts with the input.}
Observed bridge targets exceed predicted targets by $0.6\%$ for both zero reference and pure noise, but trail them by $3.1\%$ for linear reference. H10-first changes success relative to observed bridge-only constraints by $-2.4\%$, $-0.4\%$, and $+0.6\%$. The effect of H10-first selection depends on the input construction.

Predicted-interval composition averages $85.5\%$, $85.6\%$, and $87.9\%$ for zero, noise, and linear references (supplementary Table~\suppref{tab:predicted-seeds}). Linear reference ranks highest.

\subsection{Contribution of the interval branch}
\label{sec:controls}
A $2\times2$ comparison separates interval prediction from observed bridge composition (Table~\ref{tab:components}). All four variants use zero-reference inputs and readable missing keys. Adding composition to action-only supervision raises success from $78.8\%$ to $84.5\%$. Adding interval prediction alone gives $80.3\%$, while combining both gives $87.8\%$. The interval branch adds $3.3\%$ when composition is present, compared with $1.5\%$ without it.

\begin{table}[!ht]
\centering\small\setlength{\tabcolsep}{4pt}\renewcommand{\arraystretch}{1.08}
\caption{\textbf{Component isolation.} Same LIBERO 70.5\% gap manifest, zero reference, and readable missing keys. Entries report success (\%). Composition uses observed bridge endpoints. Supplementary Table~\suppref{tab:no-d} extends the with/without-D comparison to all three missing-input constructions. Reg-Read replaces interval FM with endpoint regression loss $L_R$. NoRead blocks action access to the interval token.}
\label{tab:components}
\begin{tabular*}{\columnwidth}{@{\extracolsep{\fill}}lcN@{}}\toprule
Variant & \colhead{Objective} & \colhead{Success}\\\midrule
Action FM only & $L_A$ & 78.8\\
Interval only & $L_A+L_D$ & 80.3\\
Composition only & $L_A+L_B$ & 84.5\\
Full (observed D) & $L_A+L_D+L_B$ & \bestlast{87.8}\\
Reg-Read & $L_A+L_R+L_B$ & 84.9\\
NoRead & $L_A+L_D+L_B$ & 83.1\\
\bottomrule
\end{tabular*}
\end{table}

At the same training budget, the branch-removal comparison also covers pure-noise and linear inputs (supplementary Table~\suppref{tab:no-d}). Retaining D improves success by $4.1\%$ and $0.3\%$, respectively. The smaller linear-reference difference shows that the benefit depends on the missing-input construction. D-free removes the interval token, projections, loss, and dedicated interval-start observation forward.

\paragraph{Representation controls.}
Reg-Read replaces interval FM with direct regression of normalized endpoint motion from a learned query. It retains action access to the query features and observed bridge $L_B$. NoRead retains interval FM and shared parameters while blocking action access to the interval token in every training and inference layer. Both retain the interval ranges, observation views, and the $5,000$-update, batch-$256$ training recipe. Full reaches $87.8\%$, compared with $84.9\%$ for Reg-Read and $83.1\%$ for NoRead (Table~\ref{tab:components}). The respective gains of $2.9\%$ and $4.6\%$ support the interval-FM construction and direct feature access in this zero-reference, observed-target setting.

\paragraph{Matched observation exposure.}
Exposure-matched Mask uses the main method's linear-reference inputs and readable missing keys. It replays the same ordinary starts, auxiliary interval starts, and bridge-covering action views, supervising reliable actions at those starts. Ours reaches $88.4\%$, compared with $77.9\%$ for this control (Table~\ref{tab:exposure}). The $10.5\%$ advantage supports adding interval prediction and composition when observation exposure and missing-action inputs are aligned. Supplementary Sec.~\suppref{sec:supp-recipe} gives the sampling and budget settings.

\begin{table}[!ht]
\centering\small\setlength{\tabcolsep}{4pt}\renewcommand{\arraystretch}{1.08}
\caption{\textbf{Matched observation exposure and training cost.} LIBERO, 1.2-second gaps and 70.5\% missing actions. Both models use linear-reference inputs, readable missing keys, and matched observation reads. Costs are measured training time in L20-equivalent GPU-hours. Both use 256 ordinary windows per update for 5,000 updates. Success uses 800 evaluation episodes.}
\label{tab:exposure}
\begin{tabular*}{\columnwidth}{@{\extracolsep{\fill}}lcc@{}}\toprule
Setting & Matched Mask & Ours\\\midrule
Ordinary window reads & 1,280,000 & 1,280,000\\
Additional reads & 433,654 & 433,654\\
GPU-hours & 279.36 & 279.18\\
\midrule
Success (\%) & 77.9 & \textbf{88.4}\\
\bottomrule
\end{tabular*}
\end{table}

\subsection{Longer gaps and task-stage missingness}
\begin{table}[!ht]
\centering\small\setlength{\tabcolsep}{4pt}\renewcommand{\arraystretch}{1.08}
\caption{\textbf{Gap width and missing-action coverage.} LIBERO, 20\% demonstrations, H10. Ours uses a linear reference and predicted-interval composition. Coverage is the target fraction of missing actions. Success (\%).}
\label{tab:strength}
\begin{tabular*}{\columnwidth}{@{\extracolsep{\fill}}ccNNN@{}}\toprule
Width (s) & Missing (\%) & \colhead{Mask} & \colhead{Interval only} & \colhead{Ours}\\\midrule
1.2 & 30 & 88.9 & 89.6 & \bestlast{90.6}\\
1.2 & 50 & 83.0 & 85.5 & \bestlast{89.6}\\
1.2 & 70 & 76.4 & 78.6 & \bestlast{88.4}\\
2.0 & 70 & 73.9 & 78.4 & \bestlast{84.6}\\
3.0 & 70 & 65.8 & 69.6 & \bestlast{75.4}\\
\bottomrule
\end{tabular*}
\end{table}

We vary gap coverage and duration to test the method as supervision becomes sparse and intervals extend across more fixed windows. At $1.2$-second width, Ours scores $90.6\%$, $89.6\%$, and $88.4\%$ as target missing-action coverage rises from $30\%$ to $70\%$. Mask falls from $88.9\%$ to $76.4\%$ (Table~\ref{tab:strength}). At $70\%$ coverage, increasing width from $1.2$ to $3.0$ seconds lowers both methods, but Ours retains gains of $12.0\%$, $10.8\%$, and $9.6\%$ at $1.2$, $2.0$, and $3.0$ seconds. Interval prediction alone remains closer to Mask. Supplementary Table~\suppref{tab:probes} reports the mean success of Mask and interval-only reference probes.

\begin{table}[!ht]
\centering\small\setlength{\tabcolsep}{4pt}\renewcommand{\arraystretch}{1.08}
\caption{\textbf{Fixed task-stage gaps.} LIBERO, 339 demonstrations. Ours uses pure-noise inputs and predicted-interval composition. Complete counts fully labeled demonstrations. Missing reports the action-missing fraction (\%). Other entries give success (\%).}
\label{tab:stage}
\begin{tabular*}{\columnwidth}{@{\extracolsep{\fill}}lJNNN@{}}\toprule
Condition & \colhead{Complete} & \colhead{Missing} & \colhead{Mask} & \colhead{Ours}\\\midrule
70\% bridged & 102 & 19.9 & 85.9 & \bestlast{87.6}\\
One/task & 40 & 24.8 & 79.6 & \bestlast{83.0}\\
None & 0 & 28.2 & 51.8 & \bestlast{59.9}\\
\bottomrule
\end{tabular*}
\end{table}

\paragraph{Fixed task-stage gaps.}
Unlike random gaps, these conditions remove grasp/lift labels, with two gaps for compound tasks (Table~\ref{tab:stage}). All conditions use the same $339$ demonstrations. The one-complete-per-task and no-complete conditions use packed bridge-covering chunks for composition. Predicted-interval composition with pure-noise inputs improves over Mask by $1.8\%$ when $70\%$ of demonstrations have gaps, $3.4\%$ with one complete demonstration per task, and $8.1\%$ with none. Across these conditions, complete demonstrations decrease from $102$ to $40$ to $0$, while missing-action coverage rises from $19.9\%$ to $24.8\%$ to $28.2\%$. With gaps in every demonstration, local labels and interval motion support $59.9\%$ success versus $51.8\%$ for Mask.

\subsection{Ego-informed bimanual simulation}
\begin{table}[!ht]
\centering\small\setlength{\tabcolsep}{4pt}\renewcommand{\arraystretch}{1.08}
\caption{\textbf{Ego-informed RoboTwin evaluation.} Ten tasks, 250 training demonstrations, 5,000 updates, H50 with 50 executed actions, and ten FM integration steps. Success (\%) over 800 episodes. Both composition variants use linear-reference inputs. Ours uses predicted intervals as composition targets. Clean is a fully supervised reference.}
\label{tab:robotwin-planned}
\begin{tabular}{@{}lN@{}}\toprule
Method & \colhead{Success}\\\midrule
Clean & 49.5\\
\addlinespace[2pt]
Mask & 40.3\\
Hold & 20.3\\
Linear & 38.1\\
Interval FM only & 42.8\\
Ours (observed endpoints) & 44.6\\
Ours & \bestlast{45.4}\\
\bottomrule
\end{tabular}
\end{table}

RoboTwin 2.0~\cite{robotwin2} uses ten tasks and $250$ training demonstrations with Ego-informed bimanual gaps. Models train for $5,000$ updates and predict and execute $50$ actions per chunk, using ten FM integration steps and $800$ evaluation episodes. Local-frame pose increments compose in $SE(3)$ separately for each arm. Supplementary Sec.~\suppref{sec:robotwin-transfer} and Tables~\suppref{tab:robotwin-tasks} and~\suppref{tab:transfer-match} detail gap transfer and per-arm supervision.

Ours reaches $45.4\%$, improving on Mask by $5.1\%$ and interval FM alone by $2.6\%$ (Table~\ref{tab:robotwin-planned}). Observed-endpoint composition reaches $44.6\%$. Linear and Hold reach $38.1\%$ and $20.3\%$, respectively. Clean reaches $49.5\%$, leaving a $4.1\%$ gap to complete supervision.

\subsection{Real-robot learning from labeled Ego data}
\label{sec:real}
\begin{figure*}[t]
  \centering
  \includegraphics[width=\textwidth]{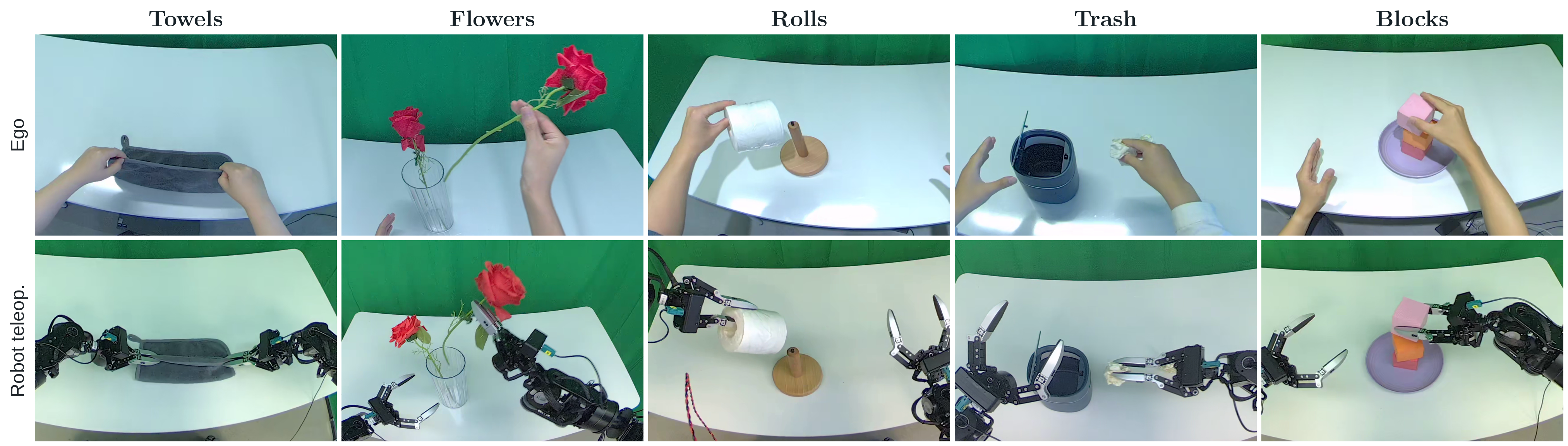}
  \caption{\textbf{Five real-world tasks.} Ego demonstrations (top) and robot teleoperation demonstrations (bottom) show towel folding, flower picking, paper-roll pickup, trash pickup, and block stacking, from left to right. Columns pair the two sources by task.}
  \label{fig:real-tasks}
\end{figure*}
\begin{table}[!ht]
\centering\small\setlength{\tabcolsep}{2pt}\renewcommand{\arraystretch}{1.08}
\caption{\textbf{Real-robot evaluation on five tasks.} Success (\%) over 50 episodes per task. Mean averages the five tasks equally. Ours uses linear-reference inputs and predicted-interval composition.}
\label{tab:real}
\sbox{\robotresultbox}{%
\begin{tabular}{@{}lNNNNNN@{}}\toprule
Method & \colhead{Towels} & \colhead{Flowers} & \colhead{Rolls} & \colhead{Trash} & \colhead{Blocks} & \multicolumn{1}{c@{}}{Mean}\\\midrule
Robot-only & 24.0 & 42.0 & 52.0 & 48.0 & 44.0 & 42.0\\
Raw & 22.0 & 26.0 & 42.0 & 38.0 & 28.0 & 31.2\\
Hold & 20.0 & 34.0 & 38.0 & 32.0 & 30.0 & 30.8\\
Linear & 30.0 & 38.0 & 60.0 & 46.0 & 38.0 & 42.4\\
Mask & 32.0 & 44.0 & 64.0 & 54.0 & 50.0 & 48.8\\
Interval FM only & 36.0 & 50.0 & 70.0 & 62.0 & 56.0 & 54.8\\
Ours & \best{42.0} & \best{54.0} & \best{78.0} & \best{66.0} & \best{58.0} & \bestlast{59.6}\\
\bottomrule
\end{tabular}%
}
\typeout{TABLE6 natural width=\the\wd\robotresultbox; max width=\the\dimexpr 0.96\linewidth\relax}
\ifdim\wd\robotresultbox>0.96\linewidth
  \resizebox{0.96\linewidth}{!}{\usebox{\robotresultbox}}%
\else
  \usebox{\robotresultbox}%
\fi
\end{table}

Figure~\ref{fig:real-tasks} shows the five tasks with examples from both training sources. Each task uses $20$ robot teleoperation demonstrations and $160$ Ego demonstrations. Mixed-data methods sample the two sources equally. Pose-label screening identifies unreliable Ego actions. Robot data provide recorded state, while missing Ego states are interpolated separately from action supervision. Deployment uses robot state. Supplementary Sec.~\suppref{sec:ego-distribution} details the labeling pipeline.

In the Ego statistics, gaps up to $0.5$\,s comprise $76.3\%$ of events but only $33.1\%$ of missing duration (supplementary Figure~\suppref{fig:ego-shares}). Gaps above $2$\,s comprise only $2.4\%$ of events but contribute $19.2\%$ of missing duration. These widths count missing-pose time, whereas supervision spans the reliable poses surrounding a complete gap. This duration imbalance motivates variable-length interval supervision alongside fixed action windows.

\paragraph{Comparison.}
We compare seven methods with $50$ evaluation episodes per task (Table~\ref{tab:real}). Robot-only uses teleoperation data. Raw retains available low-confidence Ego labels, while Hold and Linear fill flagged segments. Mask omits unreliable local supervision. Interval FM adds endpoint-motion prediction. Ours uses linear-reference inputs and constrains composed action estimates against stopped-gradient predicted interval motion. Supplementary Sec.~\suppref{sec:ego-distribution} gives the label-quality criteria, gap statistics, and additional real-robot settings.

\paragraph{Results.}
Ours reaches $59.6\%$ mean success, compared with $54.8\%$ for interval FM alone and $48.8\%$ for Mask. Gains span $2.0-8.0\%$ and $8.0-14.0\%$, respectively, across all five tasks. Raw labels yield $31.2\%$, below robot-only training at $42.0\%$. Filtering improves Ego supervision, and interval constraints add motion information.

\paragraph{Sensitivity to label screening.}
The original filter yields $59.6\%$ mean success versus $53.2\%$ for checking only the wrist and two fingertips against the hand box, a $6.4\%$ gain. Supplementary Tables~\suppref{tab:screening} and~\suppref{tab:screening-labels} give task results, label coverage, and gap statistics.

\section{Discussion and Limitations}
Endpoints constrain net motion, while reliable local labels supervise interior actions. Their quality depends on visual estimates and the mapping to robot control. Tracking failures can bias endpoints and require interpolated Ego state inputs. Contact and controller tracking errors can also separate commanded from measured motion.

\section{Related Work}
\paragraph{Generative policies and auxiliary prediction.}
Flow matching learns a velocity field along a probability path~\cite{fm}, and $\pi_{0.5}$ uses a generative action expert~\cite{pi05}. PAD jointly denoises images and actions~\cite{pad}. UMA jointly models 3D object motion and actions with available-label masks~\cite{uma}. SUAVE uses masked video/action tokens, including action-free co-training~\cite{suave}. ELAN4D supervises an action branch with robot keypoint tracks and discards its track decoder at inference~\cite{elan}. ForesightFlow predicts success potential for policy improvement and candidate ranking~\cite{fpf}. Our interval branch predicts physical endpoint motion and constrains the aggregate output of missing actions across fixed windows.

\paragraph{Incomplete action supervision.}
Behavioral cloning from observation infers actions through inverse dynamics~\cite{bco}, while latent-action pretraining learns transitions from video~\cite{lapa}. EgoMimic aligns human and robot data for co-training~\cite{egomimic}. We address label gaps after this alignment: reliable actions coexist with observed motion over missing intervals. Ambient Diffusion learns from corrupted measurements~\cite{ambient}. Flow Matching with Missing Data averages over posterior completions~\cite{missingfm}, with complete-data equivalence requiring the true completion law. Our objective instead combines local action FM, observed-interval FM, and an action-composition constraint.

\section{Conclusion}
Reliable local labels and observed endpoint motion provide complementary supervision. Our method retains action FM, constrains missing-action composition, and combines fixed windows for long gaps. Matched-observation LIBERO controls and Ego-informed simulation support this formulation, while five real-robot tasks show its value for learning from visually labeled demonstrations.
\label{sec:body_end}

{\small\bibliographystyle{ieeenat_fullname}\bibliography{references_en,ego_references}}

\begin{thebibliography}{17}
\providecommand{\natexlab}[1]{#1}
\providecommand{\url}[1]{\texttt{#1}}
\expandafter\ifx\csname urlstyle\endcsname\relax
  \providecommand{\doi}[1]{doi: #1}\else
  \providecommand{\doi}{doi: \begingroup \urlstyle{rm}\Url}\fi

\bibitem[Banerjee et~al.(2025)Banerjee, Shkodrani, Moulon, Hampali, Han, Zhang,
  Zhang, Fountain, Miller, Basol, Newcombe, Wang, Engel, and Hodan]{hot3d}
Prithviraj Banerjee, Sindi Shkodrani, Pierre Moulon, Shreyas Hampali, Shangchen
  Han, Fan Zhang, Linguang Zhang, Jade Fountain, Edward Miller, Selen Basol,
  Richard Newcombe, Robert Wang, Jakob~Julian Engel, and Tomas Hodan.
\newblock {HOT3D: Hand and Object Tracking in 3D from Egocentric Multi-View
  Videos}.
\newblock In \emph{Proceedings of the IEEE/CVF Conference on Computer Vision
  and Pattern Recognition}, pages 7061--7071, 2025.

\bibitem[Black et~al.(2025)Black, Brown, Darpinian, Dhabalia, Driess, Esmail,
  Equi, Finn, Fusai, Galliker, Ghosh, Groom, Hausman, Ichter, Jakubczak, Jones,
  Ke, LeBlanc, Levine, Li-Bell, Mothukuri, Nair, Pertsch, Ren, Shi, Smith,
  Springenberg, Stachowicz, Tanner, Vuong, Walke, Walling, Wang, Yu, and
  Zhilinsky]{pi05}
Kevin Black, Noah Brown, James Darpinian, Karan Dhabalia, Danny Driess, Adnan
  Esmail, Michael~Robert Equi, Chelsea Finn, Niccolo Fusai, Manuel~Y. Galliker,
  Dibya Ghosh, Lachy Groom, Karol Hausman, Brian Ichter, Szymon Jakubczak, Tim
  Jones, Liyiming Ke, Devin LeBlanc, Sergey Levine, Adrian Li-Bell, Mohith
  Mothukuri, Suraj Nair, Karl Pertsch, Allen~Z. Ren, Lucy~Xiaoyang Shi, Laura
  Smith, Jost~Tobias Springenberg, Kyle Stachowicz, James Tanner, Quan Vuong,
  Homer Walke, Anna Walling, Haohuan Wang, Lili Yu, and Ury Zhilinsky.
\newblock {$\pi_{0.5}$: a Vision-Language-Action Model with Open-World
  Generalization}.
\newblock In \emph{Proceedings of The 9th Conference on Robot Learning}, pages
  17--40. PMLR, 2025.

\bibitem[Cao et~al.(2026)Cao, Liu, Feng, Zhang, Lu, Owens, and Fang]{uma}
Yunhao Cao, Shitong Liu, Chao Feng, Meryl Zhang, Xuanchen Lu, Andrew Owens, and
  Kuan Fang.
\newblock {Unified Motion-Action Modeling for Heterogeneous Robot Learning}.
\newblock \emph{arXiv preprint arXiv:2606.16917}, 2026.

\bibitem[Chen et~al.(2026)Chen, Chen, Chen, Cai, Liu, Li, Liang, Lin, Ge, Gu,
  Deng, Guo, Nian, Xie, Chen, Su, Xu, Liu, Hu, Gao, Wang, Liang, Qin, Yang,
  Luo, and Mu]{robotwin2}
Tianxing Chen, Zanxin Chen, Baijun Chen, Zijian Cai, Yibin Liu, Zixuan Li,
  Qiwei Liang, Xianliang Lin, Yiheng Ge, Zhenyu Gu, Weiliang Deng, Yubin Guo,
  Tian Nian, Xuanbing Xie, Qiangyu Chen, Kailun Su, Tianling Xu, Guodong Liu,
  Mengkang Hu, Huan-Ang Gao, Kaixuan Wang, Zhixuan Liang, Yusen Qin, Xiaokang
  Yang, Ping Luo, and Yao Mu.
\newblock {RoboTwin 2.0: A Scalable Data Generator and Benchmark with Strong
  Domain Randomization for Robust Bimanual Robotic Manipulation}.
\newblock In \emph{Proceedings of the 43rd International Conference on Machine
  Learning}, pages 13673--13699. PMLR, 2026.

\bibitem[Daras et~al.(2023)Daras, Shah, Dagan, Gollakota, Dimakis, and
  Klivans]{ambient}
Giannis Daras, Kulin Shah, Yuval Dagan, Aravind Gollakota, Alex Dimakis, and
  Adam Klivans.
\newblock {Ambient Diffusion: Learning Clean Distributions from Corrupted
  Data}.
\newblock In \emph{Advances in Neural Information Processing Systems}, pages
  288--313, 2023.

\bibitem[Guo et~al.(2024)Guo, Hu, Zhang, Wang, Chen, Lu, and Chen]{pad}
Yanjiang Guo, Yucheng Hu, Jianke Zhang, Yen-Jen Wang, Xiaoyu Chen, Chaochao Lu,
  and Jianyu Chen.
\newblock {Prediction with Action: Visual Policy Learning via Joint Denoising
  Process}.
\newblock In \emph{Advances in Neural Information Processing Systems}, pages
  112386--112410, 2024.

\bibitem[He et~al.(2026)He, Yang, Fang, Zhou, Jiang, Qian, Mo, Yan, Torr, Li,
  Jiang, and Yu]{elan}
Zeyuan He, Bowen Yang, Zhirui Fang, Keru Zhou, Lei Jiang, Jingjing Qian, Fan
  Mo, Junchi Yan, Philip Torr, Xiu Li, Li Jiang, and Jialin Yu.
\newblock {ELAN4D: Embodiment-Centric 4D Supervision for Vision-Language-Action
  Models via Plug-and-Play Adaptation}.
\newblock \emph{arXiv preprint arXiv:2605.30484}, 2026.

\bibitem[Kareer et~al.(2025)Kareer, Patel, Punamiya, Mathur, Cheng, Wang,
  Hoffman, and Xu]{egomimic}
Simar Kareer, Dhruv Patel, Ryan Punamiya, Pranay Mathur, Shuo Cheng, Chen Wang,
  Judy Hoffman, and Danfei Xu.
\newblock {EgoMimic: Scaling Imitation Learning via Egocentric Video}.
\newblock In \emph{Proceedings of the IEEE International Conference on Robotics
  and Automation}, pages 13226--13233, 2025.

\bibitem[Khan et~al.(2026)Khan, Nahim, and Ju]{missingfm}
Fairoz~Nower Khan, Nabuat~Zaman Nahim, and Peizhong Ju.
\newblock {Flow Matching with Missing Data}.
\newblock \emph{arXiv preprint arXiv:2607.28698}, 2026.

\bibitem[{Lightwheel}(2026)]{lightwheel2026egosuite}
{Lightwheel}.
\newblock {EgoSuite-Open100K}: A fully annotated open egocentric human dataset.
\newblock
  \url{https://huggingface.co/collections/LightwheelAI/egosuite-open100k},
  2026.
\newblock Dataset.

\bibitem[Lipman et~al.(2023)Lipman, Chen, Ben-Hamu, Nickel, and Le]{fm}
Yaron Lipman, Ricky T.~Q. Chen, Heli Ben-Hamu, Maximilian Nickel, and Matt Le.
\newblock {Flow Matching for Generative Modeling}.
\newblock In \emph{International Conference on Learning Representations}, 2023.

\bibitem[Liu et~al.(2023)Liu, Zhu, Gao, Feng, Liu, Zhu, and Stone]{libero}
Bo Liu, Yifeng Zhu, Chongkai Gao, Yihao Feng, Qiang Liu, Yuke Zhu, and Peter
  Stone.
\newblock {LIBERO: Benchmarking Knowledge Transfer for Lifelong Robot
  Learning}.
\newblock In \emph{Advances in Neural Information Processing Systems}, pages
  44776--44791, 2023.

\bibitem[Mei et~al.(2026)Mei, He, Cao, Wang, Zhu, Zhou, Wang, Xin, Cheng, Yang,
  Wang, Zheng, Huang, Chen, and Wang]{fpf}
Yunpeng Mei, Jiakai He, Hongjie Cao, Chenyu Wang, Xiaowen Zhu, Yihan Zhou,
  Jiamin Wang, Chenbo Xin, Peng Cheng, Yuxuan Yang, Yijie Wang, Xinhu Zheng,
  Gao Huang, Jie Chen, and Gang Wang.
\newblock {Potential-Guided Flow Matching for Vision-Language-Action Policy
  Improvement}.
\newblock \emph{arXiv preprint arXiv:2606.04968}, 2026.

\bibitem[Syed et~al.(2026)Syed, Mercat, Keh, Arora, Shah, Onol, Zhang, and
  Dear]{suave}
Rhythm Syed, Jean Mercat, Sedrick Keh, Kushal Arora, Paarth Shah, Aykut Onol,
  Mengchao Zhang, and Tony Dear.
\newblock {SUAVE: Unified Video-Action Models via Masked Diffusion}.
\newblock \emph{arXiv preprint arXiv:2610.04009}, 2026.

\bibitem[Torabi et~al.(2018)Torabi, Warnell, and Stone]{bco}
Faraz Torabi, Garrett Warnell, and Peter Stone.
\newblock {Behavioral Cloning from Observation}.
\newblock In \emph{Proceedings of the Twenty-Seventh International Joint
  Conference on Artificial Intelligence}, pages 4950--4957, 2018.

\bibitem[Ye et~al.(2025)Ye, Jang, Jeon, Joo, Yang, Peng, Mandlekar, Tan, Chao,
  Lin, Liden, Lee, Gao, Zettlemoyer, Fox, and Seo]{lapa}
Seonghyeon Ye, Joel Jang, Byeongguk Jeon, Se~June Joo, Jianwei Yang, Baolin
  Peng, Ajay Mandlekar, Reuben Tan, Yu-Wei Chao, Bill~Yuchen Lin, Lars Liden,
  Kimin Lee, Jianfeng Gao, Luke Zettlemoyer, Dieter Fox, and Minjoon Seo.
\newblock {Latent Action Pretraining from Videos}.
\newblock In \emph{International Conference on Learning Representations}, pages
  28213--28239, 2025.

\bibitem[Zhang et~al.(2025)Zhang, Deng, Ma, and Potamias]{zhang2025hawor}
Jinglei Zhang, Jiankang Deng, Chao Ma, and Rolandos~Alexandros Potamias.
\newblock {HaWoR}: World-space hand motion reconstruction from egocentric
  videos.
\newblock In \emph{Proceedings of the IEEE/CVF Conference on Computer Vision
  and Pattern Recognition}, pages 1805--1815, 2025.

\end{thebibliography}

\ifdefined\SubmissionOnly\else
\clearpage
\appendix
\setcounter{figure}{0}\setcounter{table}{0}\setcounter{equation}{0}
\renewcommand{\theequation}{S\arabic{equation}}
\renewcommand{\thefigure}{S\arabic{figure}}
\renewcommand{\thetable}{S\arabic{table}}
\renewcommand{\theHfigure}{supp.\arabic{figure}}
\renewcommand{\theHtable}{supp.\arabic{table}}
\maketitlesupplementary
\section{Interval-only FM at 30\% missingness}
This study combines action and interval FM with blocked missing keys. Interval observations align with interval starts. Tables~\ref{tab:supp-setup} and~\ref{tab:supp-main} report settings and mean success.

\begin{table}[!ht]
\centering\small\setlength{\tabcolsep}{4pt}\renewcommand{\arraystretch}{1.08}
\caption{\textbf{Settings for the interval-only comparison.} ``Missing width'' counts unavailable pose samples. Endpoint span counts affected action differences.}
\label{tab:supp-setup}
\begin{tabular*}{\columnwidth}{@{\extracolsep{\fill}}lcc@{}}\toprule
Setting & Short gap & Long gap\\\midrule
Missing width (10 Hz) & 5 / 0.5 s & 12 / 1.2 s\\
Endpoint span & 6 steps & 13 steps\\
Nominal events/min & 30 & $180/13$\\
Missing action coverage & 30.0\% & 30.1\%\\
Complete bridges & 2,745 & 1,271\\
Action horizon & 10 & 10\\
Training runs & 2 & 2\\
Updates / action batch & 5,000 / 256 & 5,000 / 256\\\bottomrule
\end{tabular*}
\end{table}

\begin{table}[!ht]
\centering\small\setlength{\tabcolsep}{4pt}\renewcommand{\arraystretch}{1.08}
\caption{\textbf{Interval-only results at approximately 30\% missing actions.} Mean success (\%) over two independent training runs, with 800 evaluation episodes per model. Columns indicate gap width. Hold and Linear use filled labels. Mask and \emph{Interval FM only} both block missing action keys. All omit composition loss.}
\label{tab:supp-main}
\begin{tabular*}{\columnwidth}{@{\extracolsep{\fill}}lNN@{}}\toprule
Method & \colhead{0.5 s} & \colhead{1.2 s}\\\midrule
Hold & 80.4 & 67.2\\
Linear & 89.7 & 82.3\\
Mask & 88.8 & \bestlast{87.9}\\
Interval FM only & \best{90.0} & 87.6\\
\bottomrule
\end{tabular*}
\end{table}

Interval prediction averages 90.0\% on short gaps and 87.6\% on long gaps. Relative to Mask, the differences are $+1.2\%$ and $-0.3\%$.

\section{LIBERO implementation details}
\label{sec:supp-recipe}
All LIBERO studies use the fixed 20\% subset: 40 tasks, 339 demonstrations, and 53,712 ordinary windows. The optimizer is AdamW with moment-decay coefficients $(0.9,0.95)$, numerical stabilizer $10^{-8}$, weight decay $10^{-10}$, and gradient clipping at 1. The base recipe uses BF16, eight GPUs, two data-loader workers, 256 ordinary windows per update, and 5,000 updates. The learning-rate schedule has 10,000-step warmup, peak and decay-end rates $5\times10^{-5}$, and decay horizon $10^6$. The reported checkpoints therefore remain within warmup.

With batch size $B$ and action horizon $H$, the pretrained interface uses padded action dimension $d_a=32$. It projects $[B,H,32]$ inputs to $[B,H,1024]$ tokens. The range MLP has 128 hidden units, and the interval output projection predicts six coordinates. Auxiliary calls retain $H$ inactive action slots in the static suffix. Adding the interval token gives a suffix of $[B,H+1,1024]$, or $[B,11,1024]$ for $H=10$, while preserving pretrained tensor shapes. Backpropagation recomputes auxiliary activations.

For an interval of $L$ actions, let $\delta p_i$ denote physical translation increments and $D^p$ their net displacement. The map $D^p=\sum_{i=1}^L\delta p_i$ has rank three and nullity $3(L-1)$ over unconstrained $3L$-dimensional increments. Sequences $(1,1)$ and $(3,-1)$ have equal sums but different paths.

The composition study uses the 1.2-second gap width with nominal frequency $420/13$ events/min. Its 70.5\% missing-action manifest contains 2,977 complete bridges. The main-paper matrix compares three missing-action inputs and four composition settings. For H10-first selection, bridge groups are deduplicated within each update and capped at 32. For a bridge $[103,116)$, the grouped windows $[100,110)$ and $[110,120)$ contribute action offsets 3--9 and 0--5 to the composition. Each window uses its own starting observation. The group shares one sampled time and computation graph, with independent noise per window. Its endpoint loss contributes to one optimizer update, and extra grouped rows contribute only composition loss.

Reg-Read uses a learned query with the same range encoding and a linear six-dimensional output head. Its target uses the interval branch's normalization, and mean squared error replaces interval FM. At each of ten action sampling steps, actions can read this query. NoRead retains interval denoising and its shared-parameter gradients, while disabling action-to-interval attention throughout training and inference. Both keep zero-reference missing inputs and observed bridge composition, with interval and composition loss weights of one. Full, Reg-Read, and NoRead each use 256 ordinary action windows per update and 5,000 updates, with one paired training run per model.

The exposure control replays ordinary and extra observation starts with the same augmentation keys and records repeated reads. It combines reliable-action errors across the ordinary batch and additional windows, normalizing by valid action coordinates. Duplicate extra windows contribute supervised loss once per update. Both models use 5,000 updates with 256 ordinary windows per update, giving 1,280,000 ordinary window reads. The full sampling record gives 433,654 additional reads for Ours, which Mask matches through stepwise replay. Training costs are 279.18 L20-equivalent GPU-hours for Ours and 279.36 for Mask.

\paragraph{Command-integrated endpoints.}
For controlled experiments, we first integrate the recorded control commands into a command-space pose sequence using translation scale $0.05$ and rotation scale $0.5$. We then remove interior pose samples, keep the designated reliable endpoints, and mark affected action differences as missing. For LIBERO, let $p^{\rm cmd}_i,R^{\rm cmd}_i$ denote command-integrated position and rotation, and $c_i^p,c_i^R$ the unnormalized translation and rotation commands. Translation adds in world axes, $p^{\rm cmd}_{i+1}=p^{\rm cmd}_i+0.05c_i^p$, and rotation composes by left multiplication, $R^{\rm cmd}_{i+1}=\Exp(0.5c_i^R)R^{\rm cmd}_i$. Interval targets use the permitted command-trajectory endpoints. RoboTwin uses local-frame composition in the rigid-motion group $SE(3)$. These simulation targets encode the motion obtained by integrating recorded control commands. The real-robot study uses visually labeled Ego data.

\paragraph{Flow inputs and normalization.}
Following the base training path, draw $\xi\sim\mathrm{Beta}(1.5,1)$ and set $\tau=0.001+0.999\xi$. Superscripts $p$ and $R$ select translation and rotation components. With frozen translation-action centering and scale vectors $b_p,s_p$, normalization is
\begin{equation}
 z^p=\frac{D^p/0.05-Lb_p}{Ls_p},\qquad
 z^R=\frac{\Log(R_vR_u^\top)^\vee}{\pi},
\label{eq:norm}
\end{equation}
Division is coordinate-wise. Translation uses the action statistics and duration, while rotation uses its angular range.

\paragraph{Token embeddings and attention.}
The pretrained projection embeds each action as a token. Let $h_D$ be the interval embedding, $W_D,b_D$ its input projection and bias, and $e_D$ its learned type embedding. A multilayer perceptron (MLP) encodes the range normalized by horizon:
\begin{equation}
 h_D=W_D d_\tau + b_D + \mathrm{MLP}(q/H)+e_D.
\label{eq:embed}
\end{equation}

\paragraph{Joint sampling.}
At deployment, let $C$ be the current observation and $k$ the integration-step index. The action state $x^{(k)}$ and interval state $d_H^{(k)}$ start from independent standard Gaussian noise in normalized coordinates. Set $q=(0,H)$ and activate all action tokens and the main interval token. For $K=10$ Euler steps indexed by $k=0,\ldots,K-1$, set $\tau_k=1-k/K$ and jointly integrate
\begin{align}
 x^{(k+1)}&=x^{(k)}-K^{-1}v^A_\theta(C,x^{(k)},d^{(k)}_H,\tau_k),\\
 d^{(k+1)}_H&=d^{(k)}_H-K^{-1}v^D_\theta(C,d^{(k)}_H,q,\tau_k).
\end{align}
Action tokens read the interval hidden features at each denoising step. The controller executes the resulting action sequence. In LIBERO, inference uses ten Euler evaluations of the shared action--interval expert, matching the integration-step count of the baseline sampler.

In the three-input study, training and inference both expose all action keys. Pure-noise missing inputs follow the initial sampler distribution. During deployment, subsequent inputs evolve through the learned velocity field.

\section{Additional LIBERO comparisons}
\label{sec:additional-results}
The predicted-interval composition and D-free studies use 1.2-second gaps and 70.5\% missing actions. Training and evaluation follow the shared LIBERO protocol.

\begin{table}[!ht]
\centering\small\setlength{\tabcolsep}{4pt}\renewcommand{\arraystretch}{1.08}
\caption{\textbf{Mean success with predicted-interval composition.} LIBERO, 1.2-second gaps, 70.5\% missing actions, 5,000 updates. Mean success (\%) over three independent training runs, with 800 evaluation episodes per model.}
\label{tab:predicted-seeds}
\begin{tabular*}{\columnwidth}{@{\extracolsep{\fill}}lN@{}}\toprule
Input & \colhead{Mean success (\%)}\\\midrule
Zero reference & 85.5\\
Pure noise & 85.6\\
Linear reference & \bestlast{87.9}\\
\bottomrule
\end{tabular*}
\end{table}

Mean success is 87.9\% with linear reference, 85.5\% with zero reference, and 85.6\% with pure noise.

\begin{table}[!ht]
\centering\small\setlength{\tabcolsep}{4pt}\renewcommand{\arraystretch}{1.08}
\caption{\textbf{Interval-branch ablation with observed bridge composition.} LIBERO, 5,000 updates. Success (\%), 800 episodes per model. D-free removes the interval token, head, FM loss, and dedicated interval-view forward.}
\label{tab:no-d}
\begin{tabular*}{\columnwidth}{@{\extracolsep{\fill}}lNNN@{}}\toprule
Input & \colhead{With D} & \colhead{D-free} & \colhead{Difference}\\\midrule
Zero reference & \best{87.8} & 84.5 & +3.3\\
Pure noise & \best{87.4} & 83.3 & +4.1\\
Linear reference & \best{85.3} & 85.0 & +0.3\\
\bottomrule
\end{tabular*}
\end{table}

The interval branch improves observed-target composition for all three inputs in this comparison, by 3.3\%, 4.1\%, and 0.3\% for zero, noise, and linear reference.

\begin{table}[!htbp]
\centering\small\setlength{\tabcolsep}{4pt}\renewcommand{\arraystretch}{1.08}
\caption{\textbf{Mean success across gap settings.} Success (\%) averaged over two independent training runs, with 800 episodes per model. Widths are missing-pose durations. Missing-action fractions are targets. Ours uses predicted-interval composition.}
\label{tab:probes}
\begin{tabular*}{\columnwidth}{@{\extracolsep{\fill}}lN@{}}\toprule
Method & \colhead{Mean success (\%)}\\\midrule
\multicolumn{2}{c}{1.2-second gaps, 50\% missing actions}\\\midrule
Mask & 81.9\\
Interval FM only & 84.1\\
Ours & \bestlast{89.6}\\
Linear & 74.6\\
\midrule
\multicolumn{2}{c}{1.2-second gaps, 70\% missing actions}\\\midrule
Mask & 76.9\\
Interval FM only & \bestlast{78.0}\\
\midrule
\multicolumn{2}{c}{2.0-second gaps, 50\% missing actions}\\\midrule
Mask & 80.7\\
Interval FM only & \bestlast{80.9}\\
Linear & 58.6\\
\midrule
\multicolumn{2}{c}{2.0-second gaps, 70\% missing actions}\\\midrule
Mask & 72.6\\
Interval FM only & \bestlast{75.1}\\
\bottomrule
\end{tabular*}
\end{table}

Interval FM without composition has mixed benefits across gap regimes. At 1.2-second width, its mean success exceeds Mask by 2.3\% and 1.1\% at 50\% and 70\% missing actions. At 2.0-second width, the corresponding gains are 0.3\% and 2.5\%.

At 1.2-second width and 50\% missing actions, predicted-interval composition averages 89.6\% success, compared with 81.9\% for Mask and 84.1\% for interval FM alone. Linear averages 74.6\% at 1.2 seconds and 58.6\% at 2.0 seconds under 50\% missing actions.

\begin{figure}[!htbp]
\centering\includegraphics[width=\columnwidth]{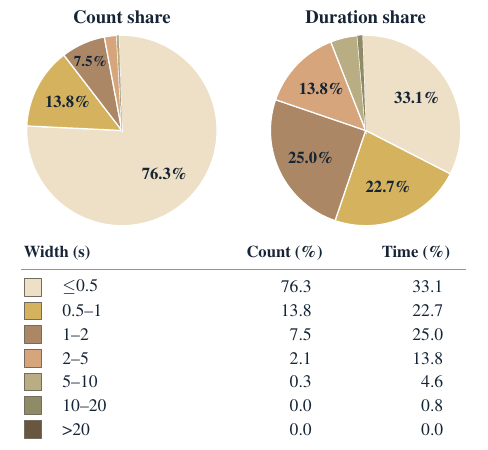}
\caption{\textbf{Real Ego gap widths, weighted by event count and missing duration.} The 45,722 segments are maximal non-green runs counted separately for each hand, including episode-boundary runs. Left: fraction of segments. Right: fraction of their summed durations. Widths sum missing-pose time cells.}
\label{fig:ego-shares}
\end{figure}

\begin{figure*}[!tp]
\centering\includegraphics[width=\textwidth]{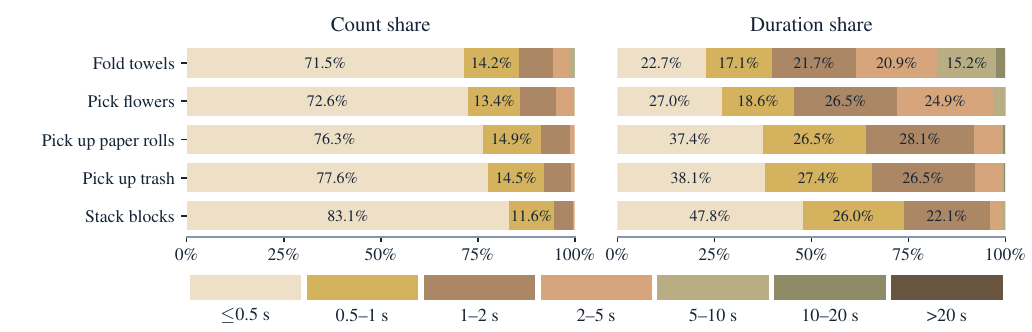}
\caption{\textbf{Task-dependent Ego gap structure.} Each task is normalized separately by its total gap count (left) or summed gap duration (right), using the width bins in Figure~\ref{fig:ego-shares}. Long gaps occupy more missing time in towel folding and flower picking than in block stacking.}
\label{fig:ego-tasks}
\end{figure*}

\begin{figure*}[!tp]
\centering\includegraphics[width=\textwidth]{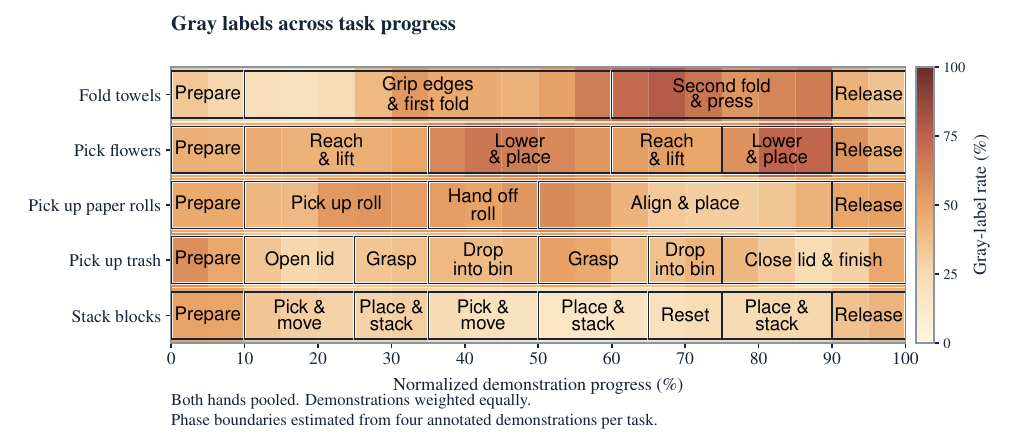}
\caption{\textbf{Label unavailability over normalized task progress and approximate action phases.} Background color shows the gray-label fraction in 5\%-progress bins, merging both hands within each demonstration and then weighting demonstrations equally. Gray denotes insufficient label evidence. The bridge plots also include other non-green states. Overlaid phase bands summarize four sampled demonstrations per task, with eight image anchors per demonstration (20 demonstrations, 160 images). Coarse phase boundaries are estimated from these sparse anchors and rounded to the 5\% grid.}
\label{fig:ego-progress}
\end{figure*}

\section{Gap structure in real Ego data}
\label{sec:ego-distribution}
HaWoR~\cite{zhang2025hawor} supplies 3D hand geometry, and a hand detector trained on Lightwheel Ego data~\cite{lightwheel2026egosuite} supplies image boxes for label screening. Wrist poses define end-effector motion. Thumb and index fingertips represent the gripper. Robot data provide recorded state. Missing Ego states are interpolated between available states. Actions use a separate reliability mask. Deployment uses robot state.

We characterize missing pose labels in 1,249 demonstrations across towel folding, flower picking, paper-roll pickup, trash pickup, and block stacking. Plate-wiping recordings are excluded because their pose tracks fail association and geometric-validity checks.

\paragraph{Many short gaps, substantial long-gap time.}
Gaps at most 0.5\,s account for 76.3\% of events but only 33.1\% of missing duration. The 1--2\,s bin contributes 7.5\% of events and 25.0\% of duration. Gaps above 2\,s contribute 2.4\% of events and 19.2\% of duration. The mean is 0.44\,s, median 0.22\,s, and maximum 18.00\,s. The 90th, 95th, and 99th percentiles (P90, P95, P99) are 1.00, 1.56, and 3.11\,s.

\paragraph{Quality filtering and duration definitions.}
Green denotes accepted pose labels, and non-green denotes insufficient evidence or geometric inconsistency. The analysis combines existing validity annotations with additional geometric checks on selected long-gap candidates. These checks use a 10-pixel reprojection threshold and detector confidence 0.60. Accepting a candidate also requires stereo support at wrist, thumb, and index points, at least 11 supported points per view, and finite, nondegenerate poses and adjacent increments. Existing validity annotations determine acceptance outside the selected candidate set.

The distribution is evaluated at sampled timestamps (approximately 0.111\,s apart). Missing duration sums the cells bounded by neighboring timestamp midpoints, clipped at episode limits. A complete interval instead spans the reliable poses before and after the run. Of 45,722 non-green runs, 43,284 are internal runs with two green endpoints. The remaining 2,438 runs touch episode boundaries. Using these endpoints for robot supervision requires valid poses expressed in the controller's coordinate convention.

\begin{table}[!ht]
\centering\small\setlength{\tabcolsep}{4pt}\renewcommand{\arraystretch}{1.08}
\caption{\textbf{Label screening in real-robot learning.} Success (\%) for the main method with the original quality filter and with three-point box screening. Both use 20 robot and 160 Ego demonstrations per task, with equal source sampling. Each condition is trained once per task and evaluated over 50 episodes per model.}
\label{tab:screening}
\begin{tabular*}{\columnwidth}{@{\extracolsep{\fill}}lNN@{}}\toprule
Task & \colhead{Original filter} & \colhead{Three-point}\\\midrule
Towels & \best{42.0} & 34.0\\
Flowers & \best{54.0} & 52.0\\
Rolls & \best{78.0} & 68.0\\
Trash & \best{66.0} & 60.0\\
Blocks & \best{58.0} & 52.0\\
\midrule
Mean & \best{59.6} & 53.2\\
\bottomrule
\end{tabular*}
\end{table}

\begin{table}[!ht]
\centering\small\setlength{\tabcolsep}{3pt}\renewcommand{\arraystretch}{1.08}
\caption{\textbf{Screening statistics on 800 Ego recordings.} A stratified quality-control sample contains 160 recordings per task. Both rules use the same sampled hand-frames. Width measures missing-pose duration within gaps bounded by accepted poses.}
\label{tab:screening-labels}
\begin{tabular*}{\columnwidth}{@{\extracolsep{\fill}}lcc@{}}\toprule
Statistic & Original filter & Three-point\\\midrule
Reliable hand-frames (\%) & 57.7 & 66.8\\
Complete gaps & 28,494 & 23,148\\
Median width (s) & 0.22 & 0.22\\
P90 width (s) & 1.00 & 0.89\\
P95 width (s) & 1.56 & 1.33\\
P99 width (s) & 3.22 & 2.78\\
\bottomrule
\end{tabular*}
\end{table}

The three-point rule requires valid 3D coordinates and image projections for the wrist, index fingertip, and thumb tip. All three projections must lie within the corresponding hand box in the original left image, including its boundary. The label-statistics comparison uses the same 800 recordings and 283,010 sampled hand-frames for both rules, with hands counted separately. Complete-gap durations use the same timestamp-cell convention as the distribution analysis. This sample is drawn from the quality-control archive described above. The original filter gives higher task success across all five tasks, while three-point screening yields more accepted labels and fewer complete gaps.

\paragraph{Real-robot hardware.}
We use a DexTeleop TeleAvatar Lite with two 7-DoF arms and DexTeleop custom parallel grippers (Figure~\ref{fig:real-robot-hardware}). A head-mounted binocular fisheye module provides a $180^\circ$ field of view and $3840\times1080$ video. Two wrist-mounted RGB-D cameras capture $1280\times720$ video. The head module and both wrist cameras acquire images at 30\,Hz. The teleoperation hardware includes a VR headset, tracking sensors, and handheld controllers.

\begin{figure}[!htbp]
\centering\includegraphics[width=\columnwidth]{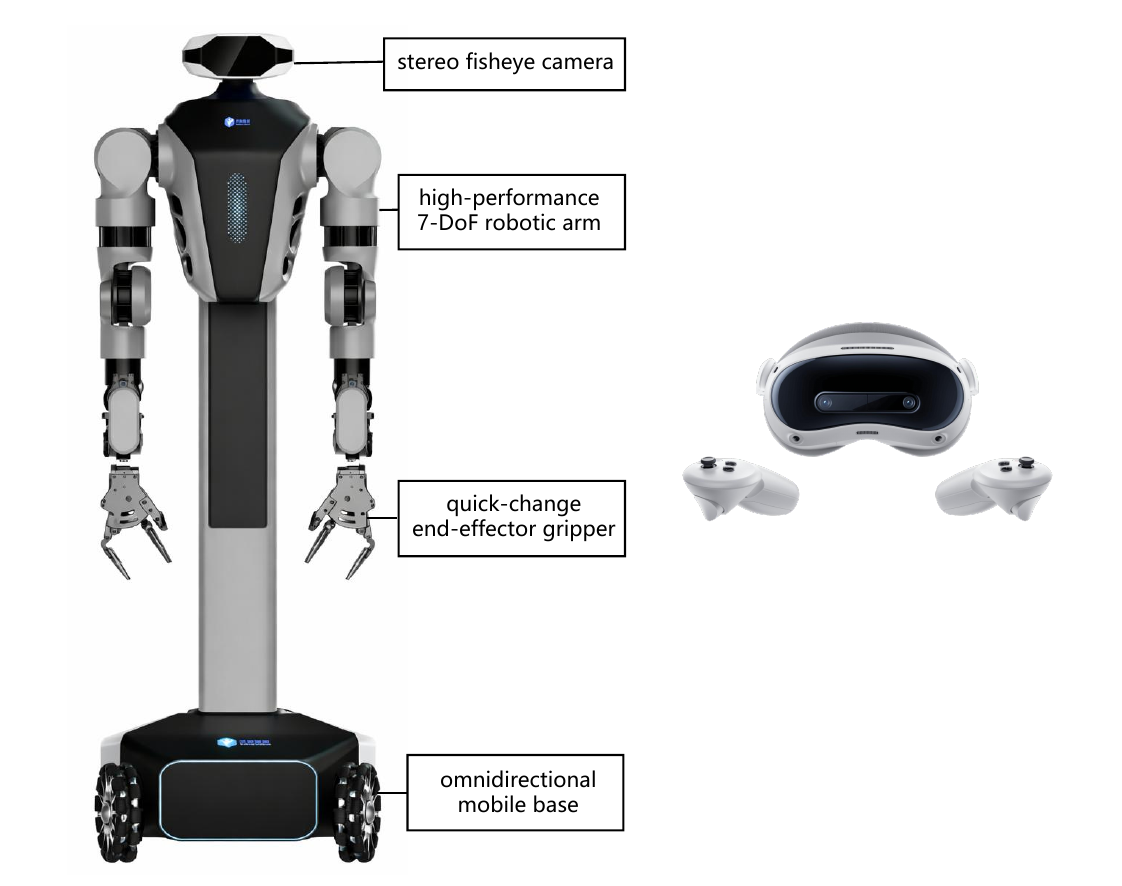}
\caption{\textbf{Real-robot platform.} The TeleAvatar Lite bimanual robot and VR teleoperation equipment used for data collection.}
\label{fig:real-robot-hardware}
\end{figure}

\paragraph{Real-robot training and control.}
We fine-tune all parameters of a $\pi_0$ policy for $20{,}000$ updates with batch size $128$, using AdamW. The cosine learning-rate schedule uses a $1{,}000$-step warmup, peak rate $2.5\times10^{-5}$, decay-end rate $2.5\times10^{-6}$, and decay horizon of $30{,}000$ steps. The policy predicts $30$ actions per chunk, and the controller runs at $20$\,Hz. Each task uses 20 robot demonstrations and 160 Ego demonstrations, sampled with equal probability from the two sources.

The wrist frame corresponds to the mounting face between the arm and gripper, and successive wrist poses provide incremental motion. Missing Ego state inputs use interpolation between available states, with spherical linear interpolation (SLERP) for rotations. Controller-coordinate mapping follows RoboTwin's end-effector-local $SE(3)$ convention. Both screening conditions use the same recordings. For each task, each condition is trained once and each model is evaluated over 50 episodes.

\paragraph{Missingness depends on task progress.}
The gray-label fraction peaks at 77.6\% around 65--70\% progress in towel folding, 74.6\% around 80--85\% in flower picking, and 59.7\% around 50--55\% in paper-roll pickup. The phase bands place these regions near the second fold/press, flower placement, and paper-roll handover, respectively. These associations motivate transferring gap patterns within corresponding task phases.

\section{Ego-informed bimanual simulation}
\label{sec:robotwin-transfer}
The RoboTwin 2.0 comparison tests Ego-derived gaps on ten tasks, using 25 training and 25 held-out demonstrations per task (250 each). Paired left/right missingness fragments are transferred between coarse corresponding phases without time stretching. Source phases use sparse image anchors. Target phases use gripper events, end-effector motion, and task structure. Pairing preserves within-fragment widths, spacings, and simultaneous loss. Cooperative grasping and held-object manipulation use coarse analogues. Episode and phase boundaries can truncate runs.

\begin{figure}[!htbp]
\centering\includegraphics[width=\columnwidth]{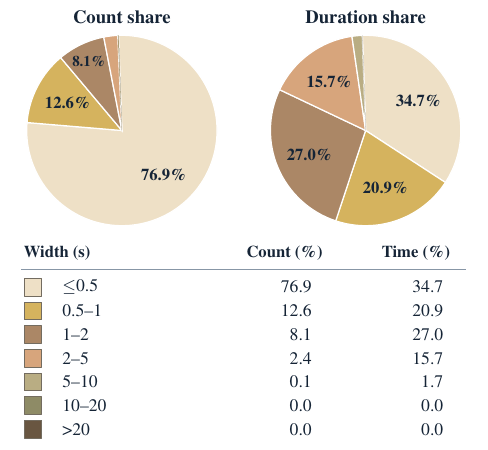}
\caption{\textbf{Ego-transferred gap widths in RoboTwin.} Count- and duration-weighted distributions of 5,767 per-arm runs in 250 training demonstrations, including boundary-open runs. Width counts missing pose samples at nominal 0.06\,s spacing, excluding reliable endpoints. Colors and bins match the real-Ego plots.}
\label{fig:robotwin-shares}
\end{figure}

\begin{figure*}[!tp]
\centering\includegraphics[width=\textwidth]{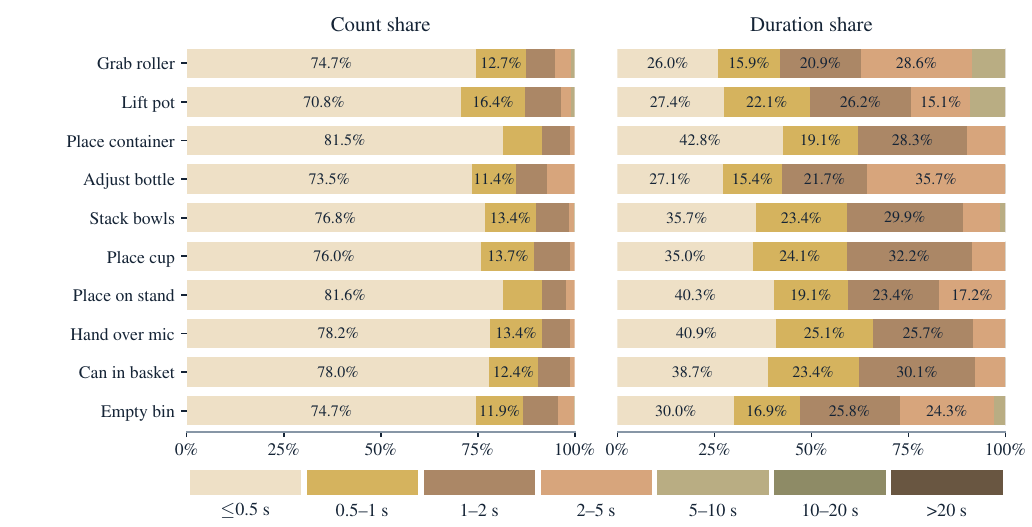}
\caption{\textbf{Task-dependent gap widths after transfer.} Each task pools both arms of 25 demonstrations and is normalized by its own gap count (left) or missing duration (right). Long gaps occupy different fractions of missing time across tasks.}
\label{fig:robotwin-tasks}
\end{figure*}

\begin{figure*}[!tp]
\centering\includegraphics[width=0.88\textwidth,trim=0 6bp 0 6bp,clip]{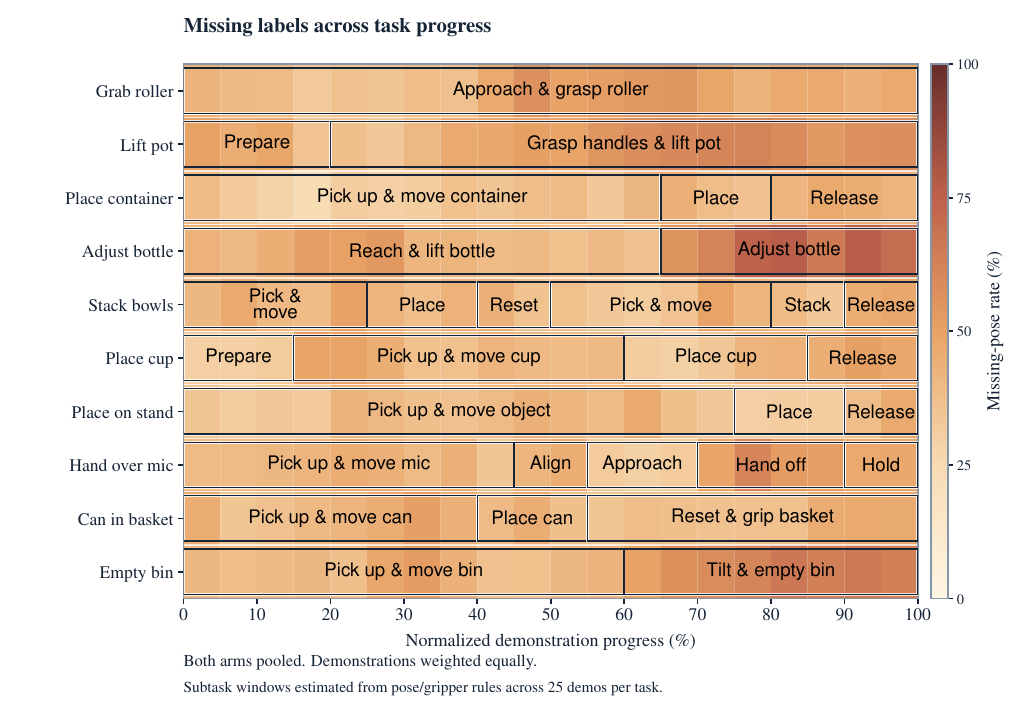}
\caption{\textbf{Transferred missingness over task progress in RoboTwin.} Color gives the missing-pose fraction in each 5\%-progress bin, pooling both arms within a demonstration and then averaging the 25 demonstrations equally. Subtask boxes summarize the majority trajectory/gripper phase across demonstrations at each bin. Short ready/reset display runs are merged into the following phase. The boxes provide coarse phase descriptions derived from trajectory and gripper rules.}
\label{fig:robotwin-progress}
\end{figure*}

\begin{table}[!ht]
\centering\small\setlength{\tabcolsep}{3pt}\renewcommand{\arraystretch}{1.08}
\caption{\textbf{RoboTwin training tasks.} Each row has 25 training demonstrations. Missing fractions (\%) are measured over all training demonstrations of that task. Only the six pose coordinates per arm are masked, while recorded robot gripper labels are retained.}
\label{tab:robotwin-tasks}
\begin{tabular*}{\columnwidth}{@{\extracolsep{\fill}}lNN@{}}\toprule
Task & \colhead{Pose} & \colhead{Action 6DoF}\\\midrule
Grab roller & 44.3 & 48.9\\
Lift pot & 50.6 & 56.2\\
Place container on plate & 36.9 & 43.0\\
Adjust bottle & 52.1 & 57.5\\
Stack two bowls & 41.0 & 47.3\\
Place cup on coaster & 39.8 & 45.6\\
Place object on stand & 37.7 & 43.5\\
Handover microphone & 43.0 & 49.7\\
Place can in basket & 41.2 & 47.6\\
Dump small bin & 48.6 & 54.3\\\bottomrule
\end{tabular*}
\end{table}

\begin{table}[!ht]
\centering\small\setlength{\tabcolsep}{3pt}\renewcommand{\arraystretch}{1.08}
\caption{\textbf{Ego and transferred RoboTwin gap distributions.} Both distributions use a common nominal 0.06\,s grid, with Ego statistics recomputed on that grid.}
\label{tab:transfer-match}
\begin{tabular*}{\columnwidth}{@{\extracolsep{\fill}}lTT@{}}\toprule
Statistic & \colhead{Ego} & \colhead{RoboTwin}\\\midrule
Missing pose fraction (\%) & 42.7 & 43.1\\
Both hands missing (\%) & 23.1 & 23.2\\
Events / hand-minute & 58.2 & 62.1\\
Mean width (s) & 0.44 & 0.42\\
Median / P90 (s) & \multicolumn{1}{c}{0.24 / 1.02} & \multicolumn{1}{c}{0.24 / 1.02}\\
P95 / P99 (s) & \multicolumn{1}{c}{1.56 / 3.12} & \multicolumn{1}{c}{1.56 / 2.76}\\
Maximum (s) & 18.00 & 5.52\\
Missing time in gaps $>5$\,s (\%) & 5.4 & 1.7\\\bottomrule
\end{tabular*}
\end{table}

\paragraph{Transferred gap distribution.}
There are 5,767 per-arm gaps and 49.1\% missing 6DoF actions. Coverage and central width quantiles are close to the source. Gaps longer than 5\,s account for 1.7\% of transferred missing time, compared with 5.4\% in the source (Table~\ref{tab:transfer-match}).

\paragraph{Bimanual labels and model.}
Actions contain two six-dimensional pose increments and two gripper commands, with per-coordinate, per-arm masks. A missing pose invalidates both adjacent differences. The bimanual suffix has 50 joint-action tokens and two six-dimensional interval tokens with side identifiers and a shared head. Each interval reads its starting observation without access to action keys. Actions can read main-interval tokens. Bridge targets and composition are arm-specific.

At nominal 0.06\,s spacing, the 50-action horizon (H50) spans 3\,s. Of 5,295 complete bridges, 30 require two bridge-aligned windows. The 472 boundary-open gaps have no interval target. Valid endpoints exist for 34.7\% of these per-arm interval slots. Grouped actions share one computation graph and one optimizer update. The other arm's reliable labels remain active. Each method uses 5,000 updates and 800 evaluation episodes.

\paragraph{Closed-loop comparisons.}
Clean restores labels on the same training demonstrations. Other methods share one bridge manifest. Hold/Linear use completed poses for FM inputs and targets, including rotation interpolation for Linear. Mask and interval-only models block missing action keys. The composition model uses readable keys and linear-reference missing inputs. Composition targets are stopped-gradient interval predictions. A separate control uses observed endpoint motion. Initialization, ordinary sampling, controller, updates, and evaluation conditions are common. Interval and grouped-action views add observations to the ordinary action batch.

\ifdefined\StandaloneSupplement
\begin{table}[!ht]
\centering\small\setlength{\tabcolsep}{4pt}\renewcommand{\arraystretch}{1.08}
\caption{\textbf{Ego-informed RoboTwin evaluation.} Ten tasks, 250 training demonstrations, 5,000 updates, H50 with 50 executed actions, and ten FM integration steps. Success (\%) over 800 episodes. Both composition variants use linear-reference inputs. Ours uses predicted intervals as composition targets. Clean is a fully supervised reference.}
\label{tab:robotwin-planned}
\begin{tabular}{@{}lN@{}}\toprule
Method & \colhead{Success}\\\midrule
Clean & 49.5\\
\addlinespace[2pt]
Mask & 40.3\\
Hold & 20.3\\
Linear & 38.1\\
Interval FM only & 42.8\\
Ours (observed endpoints) & 44.6\\
Ours & \bestlast{45.4}\\
\bottomrule
\end{tabular}
\end{table}

\else
Table~\ref{tab:robotwin-planned} compares all seven methods.
\fi

\paragraph{Transferred widths and task variation.}
Gaps up to 0.5\,s account for 76.9\% of transferred events and 34.7\% of missing duration (Figure~\ref{fig:robotwin-shares}), versus 76.3\% and 33.1\% in native-time Ego data. The bottle-adjustment 2--5\,s bin contains 7.2\% of events and 35.7\% of missing duration (Figure~\ref{fig:robotwin-tasks}).

Gaps follow task phases (Figure~\ref{fig:robotwin-progress}).

\begin{figure*}[!tp]
\section{Bimanual pose-label gaps}
Figure~1(a) in the main paper uses towel folding to illustrate how observations remain available while supervision differs between hands. Figure~\ref{fig:ego-task-examples} extends this illustration to four additional tasks, linking each selected image to its state on the left- and right-hand tracks.

\centering
\includegraphics[width=.89\textwidth]{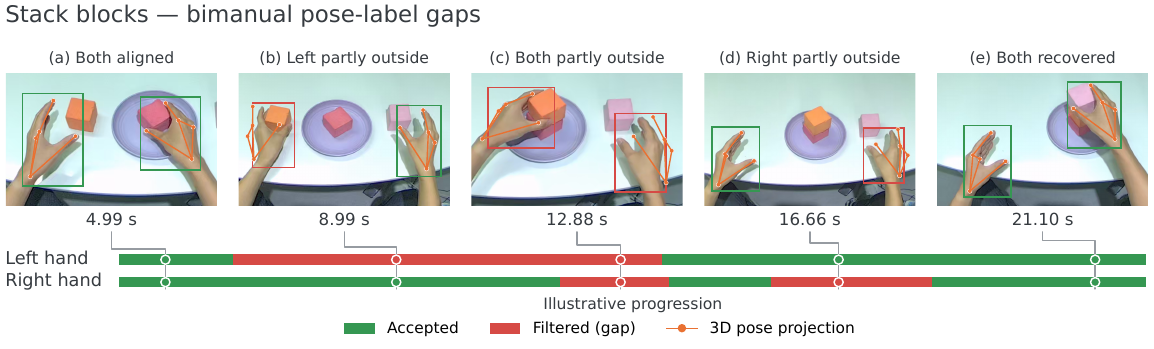}\par
\includegraphics[width=.89\textwidth]{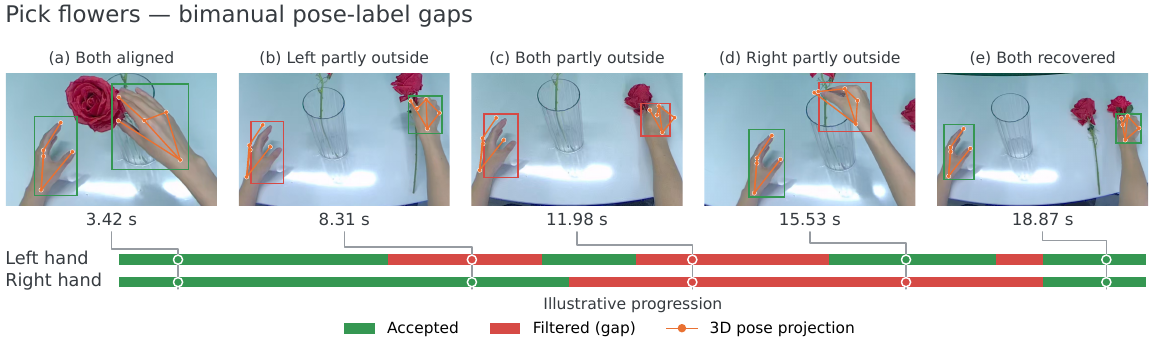}\par
\includegraphics[width=.89\textwidth]{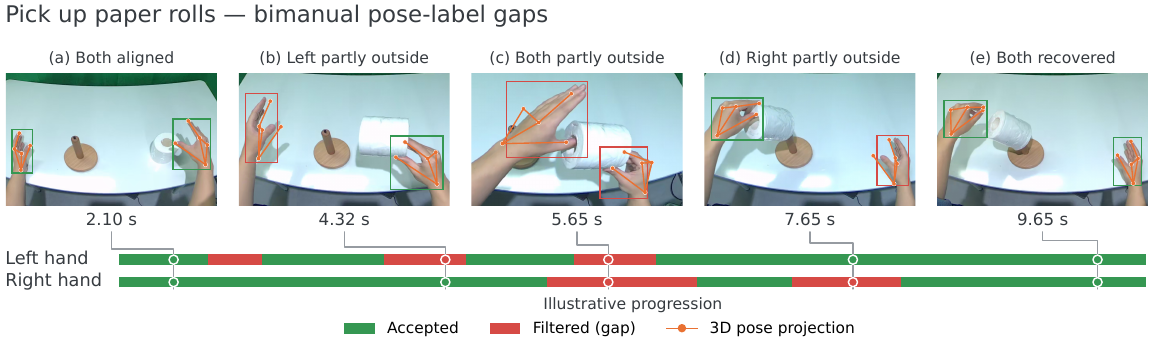}\par
\includegraphics[width=.89\textwidth]{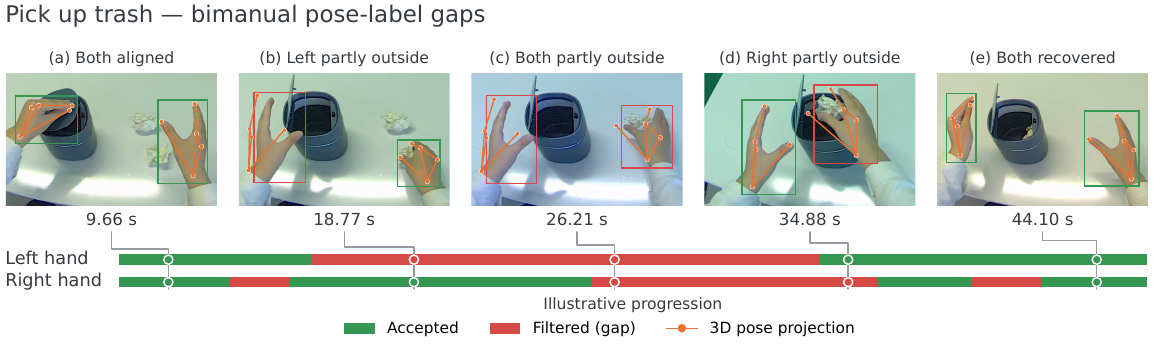}
\caption{\textbf{Bimanual gap illustrations across four Ego tasks.} From top to bottom: block stacking, flower picking, paper-roll pickup, and trash pickup. Real frames retain matched hand boxes and five projected 3D landmarks per hand. Red examples apply small, constructed pose-label offsets. The paired tracks use task-specific illustrative intervals. Each selected image matches its displayed left/right state.}
\label{fig:ego-task-examples}
\end{figure*}

\fi
\end{document}